\documentclass[sigconf]{acmart}

\usepackage{booktabs} 
\usepackage{multirow}

\copyrightyear{2019}
\acmYear{2019} 
\setcopyright{iw3c2w3}
\acmConference[WWW '19 Companion]{Companion Proceedings of the 2019 World Wide Web Conference}{May 13--17, 2019}{San Francisco, CA, USA}
\acmBooktitle{Companion Proceedings of the 2019 World Wide Web Conference (WWW '19 Companion), May 13--17, 2019, San Francisco, CA, USA}
\acmPrice{}
\acmDOI{10.1145/3308560.3316739}
\acmISBN{978-1-4503-6675-5/19/05}

\usepackage{float}
\usepackage{enumerate}
\usepackage{adjustbox}
\usepackage{subfigure}
\usepackage{xspace}
\usepackage[11pt]{moresize}
\usepackage{balance}

\newcommand{\ourdata}{PoliticalNews\xspace}
\newcommand{\uselectiondata}{US-Election2016\xspace}
\newcommand{\michiganapproach}{Fake News Detector\xspace}
\newcommand{\michiganapproachabbrev}{FNDetector\xspace}
\newcommand{\dataset}{dataset\xspace}
\newcommand{\datasets}{datasets\xspace}
\newcommand{\Dataset}{Dataset\xspace}

\renewcommand{\paragraph}[1]{\vspace{0.1cm}\noindent \textbf{#1}}

\begin{document}

\title{A Topic-Agnostic Approach for Identifying Fake News Pages}

\author{Sonia Castelo}
\affiliation{%
	\institution{New York University}
}
\email{s.castelo@nyu.edu}

\author{Thais Almeida}
\affiliation{%
	\institution{Federal University of Amazonas}
}
\email{tga@icomp.ufam.edu.br}

\author{Anas Elghafari}
\affiliation{
	\institution{New York University}
}
\email{anas.elghafari@nyu.edu}

\author{A\'ecio Santos}
\affiliation{
	\institution{New York University}
}
\email{aecio.santos@nyu.edu}

\author{Kien Pham}
\affiliation{
	\institution{New York University}
}
\email{kien.pham@nyu.edu}

\author{Eduardo Nakamura}
\affiliation{
	\institution{Federal University of Amazonas}
}
\email{nakamura@icomp.ufam.edu.br}

\author{Juliana	Freire}
\affiliation{%
	\institution{New York University}
}
\email{juliana.freire@nyu.edu}

\renewcommand{\shortauthors}{Castelo, et al.}

\begin{abstract}
Fake news and misinformation have been increasingly used to
manipulate popular opinion and influence political processes. To
better understand fake news, how they are propagated, and how to
counter their effect, it is necessary to first identify
them. Recently, approaches have been proposed to automatically
classify articles as fake based on their content. An
important challenge for these approaches comes from the dynamic
nature of news: as new
political events are covered, topics and discourse
constantly change and thus, a classifier trained
using content from articles published at a given time is likely to
become ineffective in the future. To address this challenge, we
propose a topic-agnostic (TAG) classification strategy that
uses linguistic and web-markup features to identify fake news pages.
We report experimental results using multiple data sets which show
that our approach attains high accuracy in the
identification of fake news, even as topics evolve over time.
\end{abstract}



\maketitle

\section{Introduction}
Fake news have been increasingly used to manipulate public opinion and influence political processes.
This has been made possible both by the existing Web infrastructure and online media platforms (e.g., Facebook, Twitter), which make it easy for information to be propagated. Unlike traditional print media, information can be published on Web sites and shared among users in social media platforms with no third party filtering or fact-checking~\cite{allcott2017social}.
Given that 62\% of adults in the US consume news from social media~\cite{pew2016} and many who see fake stories report that they believe them~\cite{silverman2016},  these platforms have become a target for propaganda campaigns~\cite{bakir2018fake, lazer2018science}. 

While fake news have attracted substantial attention, the problem is not well understood~\cite{lazer2018science}.
It is challenging to discover and cross-reference conflicting sources and claims. This problem is compounded due to the large number of news sites and high volume of content.
Automated methods that identify potential fake news and unreliable news sources can aid manual fact checking by providing contextual information and limiting the volume of content that the human fact-checker needs to consider. Such methods can also help us better understand the ecosystem of fake news: where they start, how they propagate, and how to counter their effects.

However, the automatic identification of fake news is a hard problem, given that news cover a wide
variety of topics and linguistic styles, and can be shared on many different platforms~\cite{shu2017fake}.

In this paper, we study the problem of detecting fake news published on the Web. Given a web page $P$, our goal is to determine whether $P$ is likely to contain fake news. Since some sites publish a mix of fake and real news~\cite{allcott2017social}, we consider pages published by suspicious sites as unreliable, and pages published by legitimate media outlets as reliable. 
We note that there is no widely-accepted definition for fake news.  Here, we focus on all types of active political misinformation that go beyond old-fashioned partisan bias, and consider unreliable not only sites that publish fabricated stories, but also sites that have a pattern of misleading headlines, thinly-sourced claims, and that promote conspiracy theories. We exclude from our definition satire sites as well as opinion sites -- even if extreme -- if they do not display a pattern of promoting misinformation.

Previous approaches to fake news identification have largely focused on using the content of news web pages to determine their veracity~\cite{rashkin2017truth,perez@coling2018}. However, using the content has important limitations. Notably, given the dynamic nature of news, as new events are covered, topics and discourse constantly change and thus, a classifier trained using content from articles published at a given time is likely to become ineffective in the future.
In addition, studies have found that page content alone is not sufficient to accurately classify the veracity of news~\cite{fairbanks2018credibility,wu2018tracing, barron2019}.

\paragraph{Our Approach.} 
To address this limitation, we \emph{propose a new classification strategy that is topic-agnostic}. 
Instead of using the bag of words in a page, we explore topic-agnostic features, including web-markup and linguistic features that are common in fake news. 

We perform a \emph{detailed experimental evaluation} using publicly-available \datasets~\cite{allcott2017social, perez@coling2018} and a new dataset we created -- the \ourdata \dataset. Since existing datasets contain a small number of articles or cover a narrow time span, they are insufficient to assess the topic-agnosticism aspect of our approach. 
The \ourdata \dataset contains a mix of political topics and spans several years. 
We report results which show that our approach is effective, obtaining accuracies that are between 8\% and 24\% higher than the baseline, and it is robust as topics change over time as well when applied to different domains. 

\paragraph{Contributions.} 
Our main contributions are:
(i) we propose a classification strategy for identifying fake news which, to the best of our knowledge, is the first to rely solely on topic-agnostic features; and (ii) we present the results of an experimental evaluation, using multiple \datasets and considering various baselines, which show that our approach is robust and attains high accuracy.

\section{Related Work}
\label{sec:related}
In this section, we discuss techniques that have been proposed to detect fake news published on Web sites. We also discuss publicly-available \datasets that have been used to evaluate these techniques.

\paragraph{Detecting Fake News on the Web.}
\citet{potthast@arxiv2017} investigated the use of writing style which included 
features such as n-grams and readability scores. They found that while style-based and topic-based classifiers are effective at differentiating hyper-partisan news from mainstream news (0.75 accuracy), they are not effective at differentiating fake from real news (0.55 accuracy).
~\citet{horne2017just} used linguistic features including readability scores, sentence structure, and part of speech of the words used. These features were very effective for the task of differentiating satire from real news (0.91 accuracy), but somewhat less so for differentiating fake news from real (0.78 accuracy). 
~\citet{perez@coling2018} also considered writing style and proposed the use of features that capture content-based aspects of web pages, such as n-grams, punctuation,  psycho-linguistic, readability and syntax. Their model attains accuracy up to 0.76. 
~\citet{fairbanks2018credibility} investigated whether credibility and bias can be assessed using content-based and structure-based methods. The structure-based method constructs a reputation graph where each node represents a site, and the edges represent mutually linked sites, as well as shared files. 
This work shares some elements with our work in its usage of features derived from the HTML source of the pages (they use a subset of the feature in our classifier), but their focus is on the network between sites.

The picture that emerges from these approaches is that content-based features, while effective for detection of bias and satire, often fall short for detecting fake news. Some of works that have achieved good results on fake news detection have done so by including additional information about the sites. In our approach, we follow a similar direction and examine the combination of linguistic 
style with sites appearance. 
Because these features do not rely on the actual content of pages, our approach is topic-agnostic and robust; this is in contrast to models that use content (e.g., n-grams), which must be retrained as the topics in the news shift.

\paragraph{Fake News Corpora.}
Despite the recent research efforts focusing on fake news, there is a dearth of publicly-available \datasets focusing on \emph{web content}. Some of the public \datasets relevant to fake news detection are: BS Detector\footnote{https://www.kaggle.com/mrisdal/fake-news/home}, BuzzFeed~\cite{buzzfeed2017git}, FakeNewsNet~\cite{shu2017exploiting},  NewsMediaSources~\cite{D18-1389}, \uselectiondata~\cite{allcott2017social} and Celebrity~\cite{perez@coling2018}. These datasets are limited with respect to size, the time period they cover, and variety of topics covered.
Since we aim to determine the time-invariant and topic-invariant features of political fake news on the Web, these \datasets are insufficient to evaluate our work.

\begin{figure}[t]
	\centering
	\subfigure[Fake News: Clash Daily]{
		\label{fn_screenshots_cd}
		\includegraphics[width=25mm,,height=30.8mm]{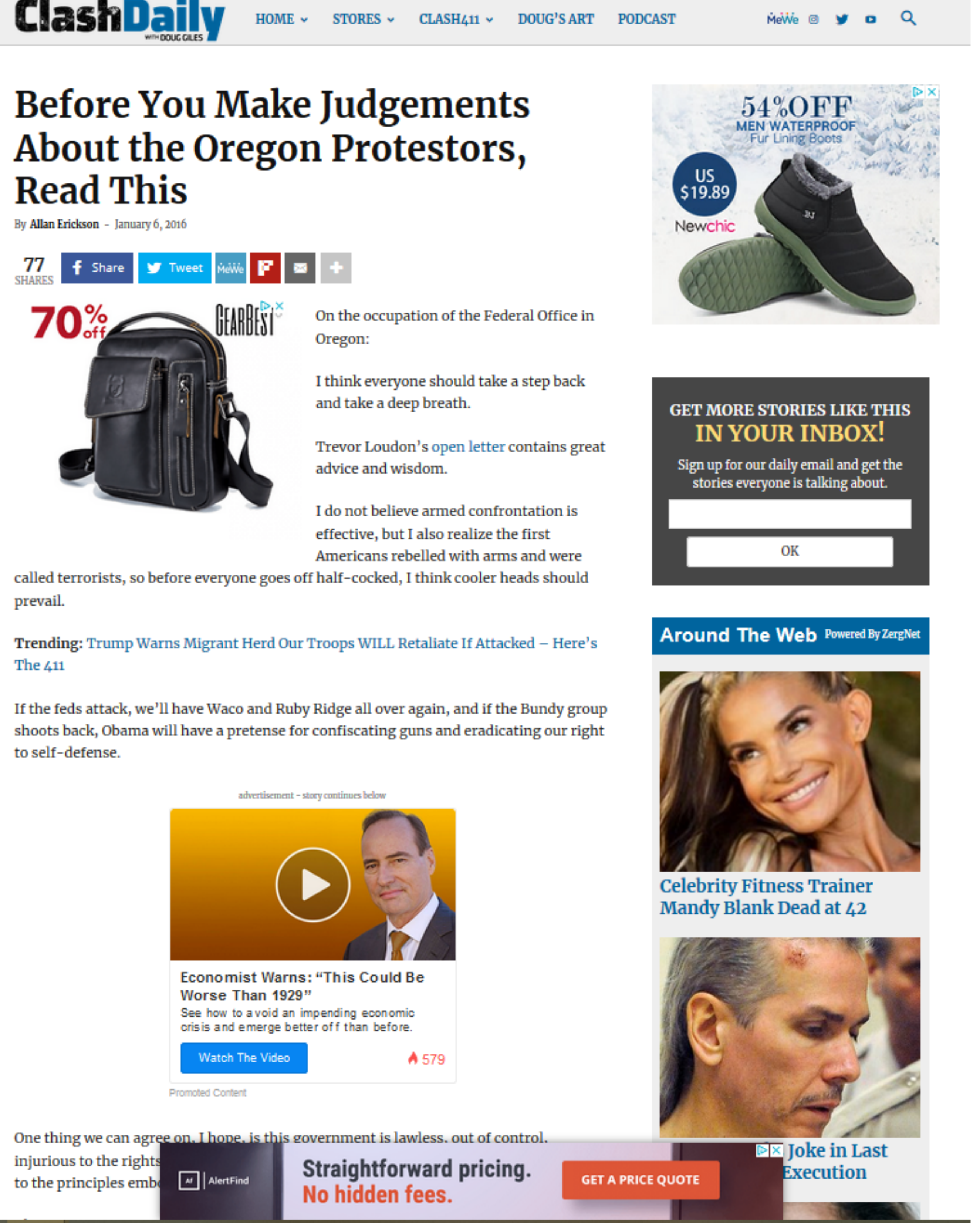}
	}\hspace{0mm}
	\subfigure[Fake News: CDP]{
		\label{fn_screenshots_CDP}
		\includegraphics[width=27.5mm,height=30.8mm]{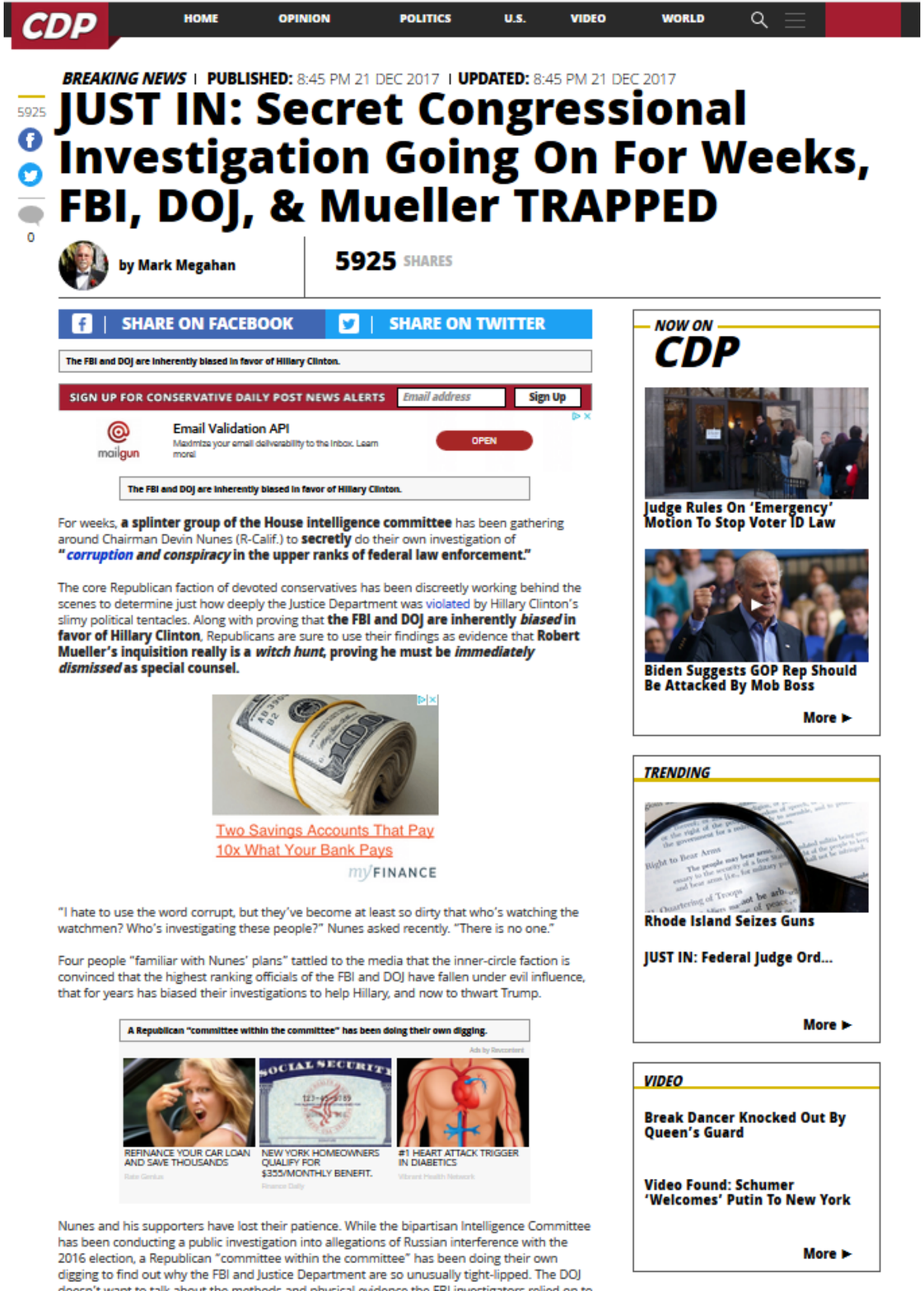}
	}\hspace{0mm}
	\subfigure[Real News: Reuters]{
		\label{msm_screenshots_r}
		\includegraphics[width=25mm,height=30.8mm]{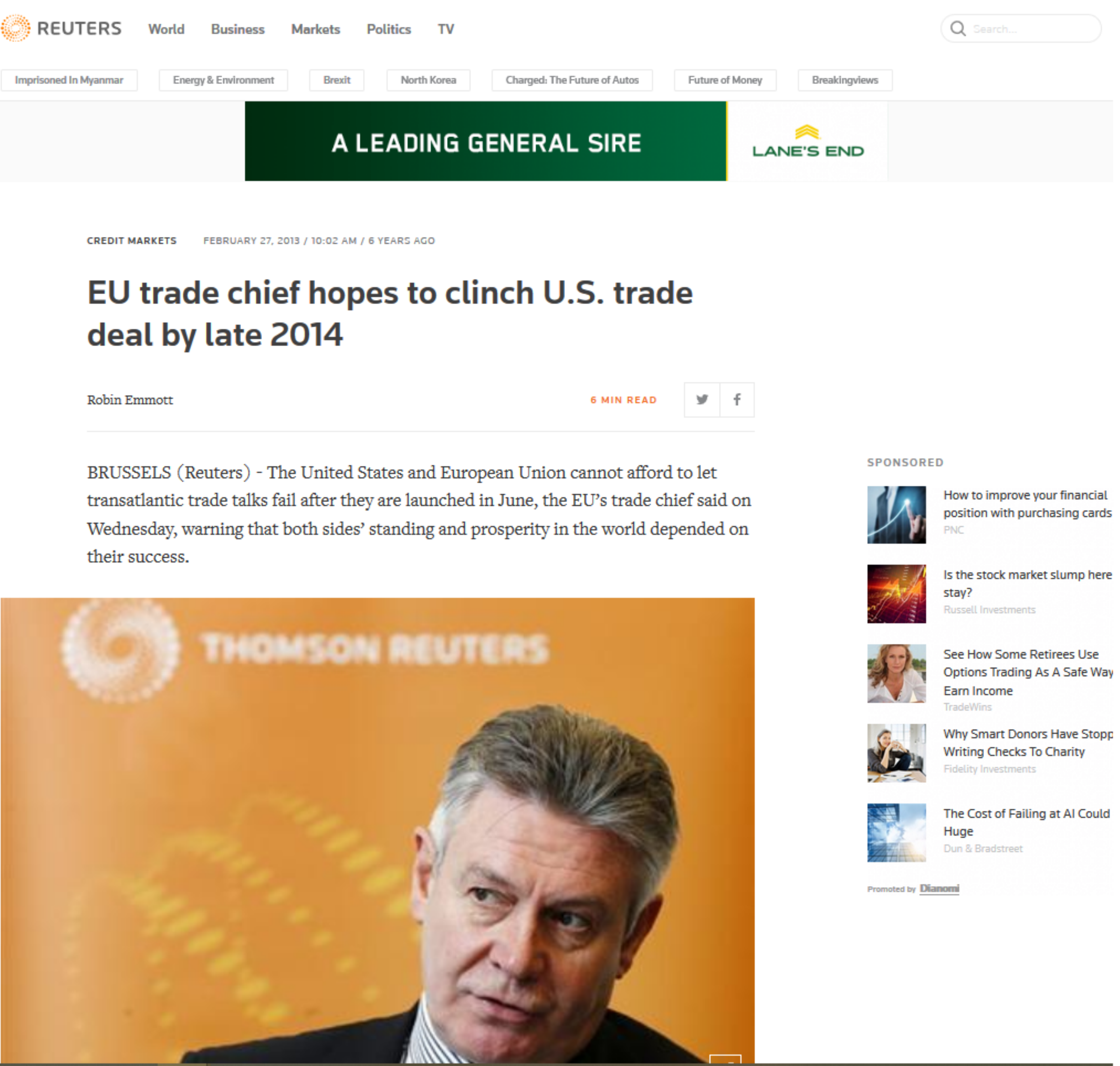}
	}
	\vspace{-10pt}
	\caption{Web pages from unreliable and reliable new sites.}
	\vspace{-10pt}
	\label{fig:fn_screenshots}
\end{figure}

\section{Topic-Agnostic Classification}
\label{sec:method}

In this section, we present our approach to fake news classification.

\begin{table*}[t]
	\centering
	\footnotesize
	\caption{Linguistic and web-markup features used to represent news articles.}
	\vspace{-9pt}
	\begin{adjustbox}{width=1\textwidth}
		
		\begin{tabular}{cllllllllll}
			\toprule
			\multirow{10}{*}{\begin{tabular}[c]{@{}c@{}}Morphological \\ Features\end{tabular}} & \textbf{Abbr.} & \textbf{Description}     & \textbf{Abbr.} & \textbf{Description}       & \textbf{Abbr.} & \textbf{Description}       & \textbf{Abbr.} & \textbf{Description}     & \textbf{Abbr.} & \textbf{Description}  \\
			
			& WDT  & WH-determiner      &PDT & Pre-determiner                          &JJ                              & Adjective or numeral, ordinal                 &VB                              & Verb, base form  & MD   & Modal auxiliary \\
			& CD     & Numeral, cardinal   &VBD & Verb, past tense                       &VBG                             & Verb, present participle or gerund       &VBN                             & Verb, past participle  & RP    & Particle \\
			& DT     & Determiner             &NNPS & Noun, proper, plural              &NN                              & Noun, common, singular or mass     &CC                              & Conjunction, coordinating  &WRB    & Wh-adverb \\
			& FOW  & Foreign word         &NNS  & Noun, common, plural              &TO                              & "To" as preposition/infinitive, superlative     					    &WP\$                            & WH-pronoun, possessive  & JJS   & Adjective, superlative\\
			& WP    & WH-pronoun         &POS   & Genitive marker                       &VBP                             & Verb, present tense, not 3rd singular                 &RBR     & Adverb, comparative  & NNP & Noun, proper, singular\\
			&UH     & Interjection            &PRP    & Pronoun, personal                 &VBZ                             & Verb, present tense, 3rd singular             &RBS     & Adverb  superlative  &PRP\$  & Pronoun, possessive \\
			&RB       & Adverb         & JJR   & Adjective, comparative  &IN  & Preposition or conjunction, subordinating   &&\\
			\midrule
			
			\multirow{4}{*}{\begin{tabular}[c]{@{}c@{}}Psychological\\   Features\end{tabular}}      &SO  &Social (family, friend)     &SD  &Summary Dimensions         & BP &Biological Processes (ingest, health, sexual)                   &AF  &Affect (anger, sad, anxiety)  &PC  &Personal Concerns     \\
			&RL  & Relativity (space, time) & FW   & Function Words   &TR   &Time Orientation (focuspast, focuspresent)    & PP    & Perceptual Process& IL   & Informal Language  \\
			&DR    & Drives (power, risk)& PM   & Punctuation Marks   &OG     &Other Grammar (quantifiers, interrogatives) & CP    & Cognitive Processes &&    \\
			
			\midrule
			\multirow{5}{*}{\begin{tabular}[c]{@{}c@{}}Readability\\   Features\end{tabular}} & FRI                              &      Flesch Reading Ease                                & WS                              & Words per sentence                  & LW                              & Long words      &                LWI              &       Linsear Write    & CLI                             &     Coleman-Liau   \\
			&FKI                              &     Flesch Kincaid Grade             & CW                             & Capitalized words       & SY                              & Syllables    &    CW    &           Complex words   & DW                             & Difficult words \\
			& MSI                             &   McLaughlin’s SMOG                  & LX                              & Lexicon  & PS                             &  Percentage of stop words     &   ARI             &    Automated Readability  &  W         &      Words      \\
			& GFI                       &  Gunning Fog   & URL                              & URLs                    & STC                             & Sentences          &      CH       &   Characters                   \\
			
			\midrule
			\multirow{4}{*}{\begin{tabular}[c]{@{}c@{}}Web-markup\\   features\end{tabular}} & AU                              & Author                                    & IT                              & Images (e.g., img, canvas)                  & ST                              & Semantics (e.g., article, section)     & FRT                             & Frames (e.g., frame, frameset)     & LT                              & Lists (e.g., ul, ol, li)        \\
			& BT                              & Basic (e.g., title,h1, p)           & FT                              & Formatting (e.g., acronym)      & FIT                             & Forms and inputs (e.g., textarea, button)    & MT                              & Metainfo (e.g., head, meta)         &  TT                              & Tables (e.g., tbody, tfoot)           \\
			& AVT                             & Audio and video     & LKT                             & Links (e.g., a, nav, link)                  & PT                              & Programming (e.g., script, object)       & ADS &  Advertisements   \\
			\bottomrule
		\end{tabular}
	\end{adjustbox}
	\label{table:topic_agnostic_features}
\end{table*}

\begin{figure*}[t]
	\subfigure[Morphological]{
		\label{fig:top15_dist_nltk}
		\includegraphics[width=45mm,,height=47.1mm]{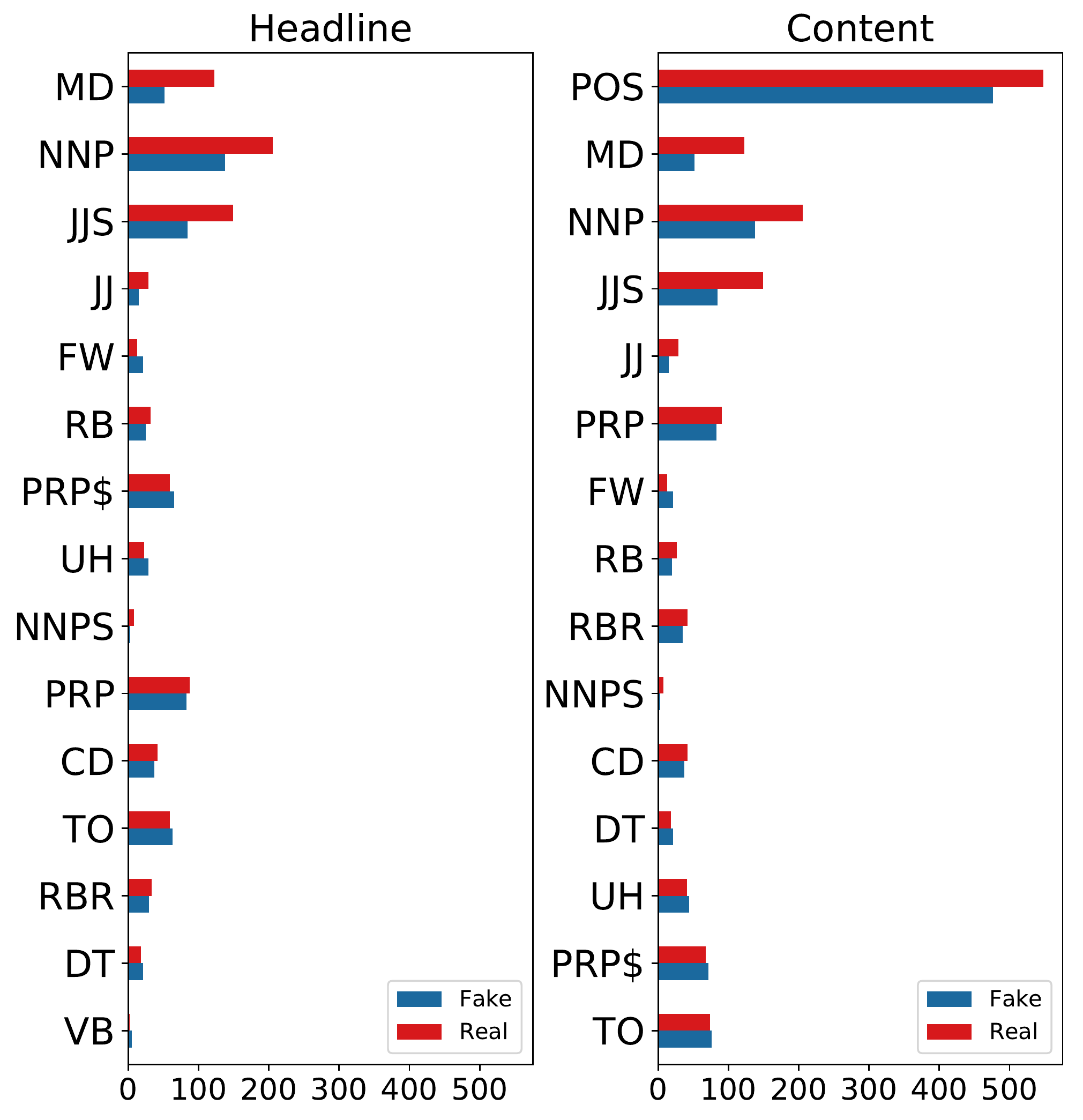}
	}
	\subfigure[Psychological]{
		\label{fig:top15_dist_liwc}
		\includegraphics[width=45mm,height=47.1mm]{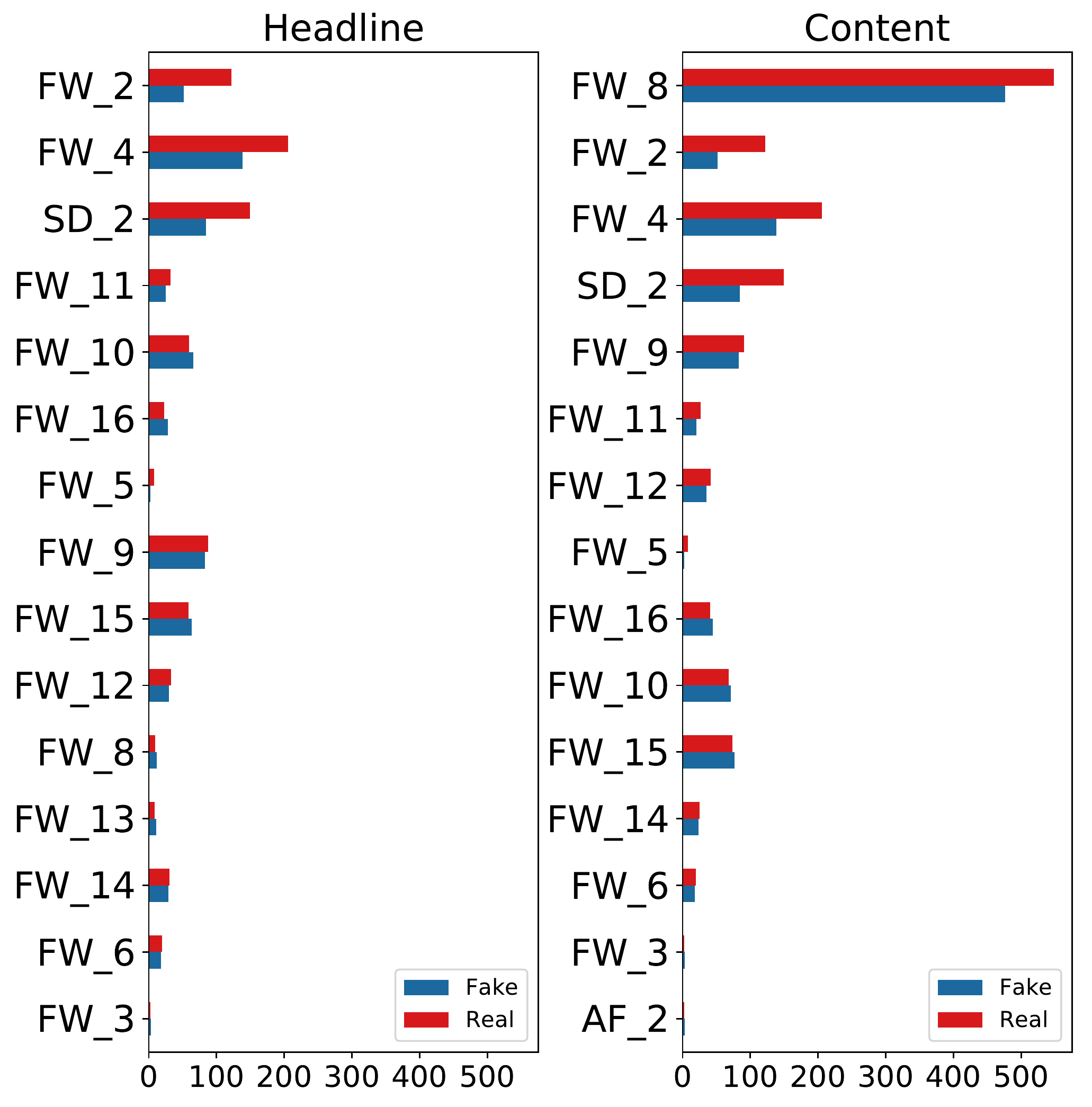}
	}
	\subfigure[Readability]{
		\label{fig:top15_dist_readby}
		\includegraphics[width=45mm]{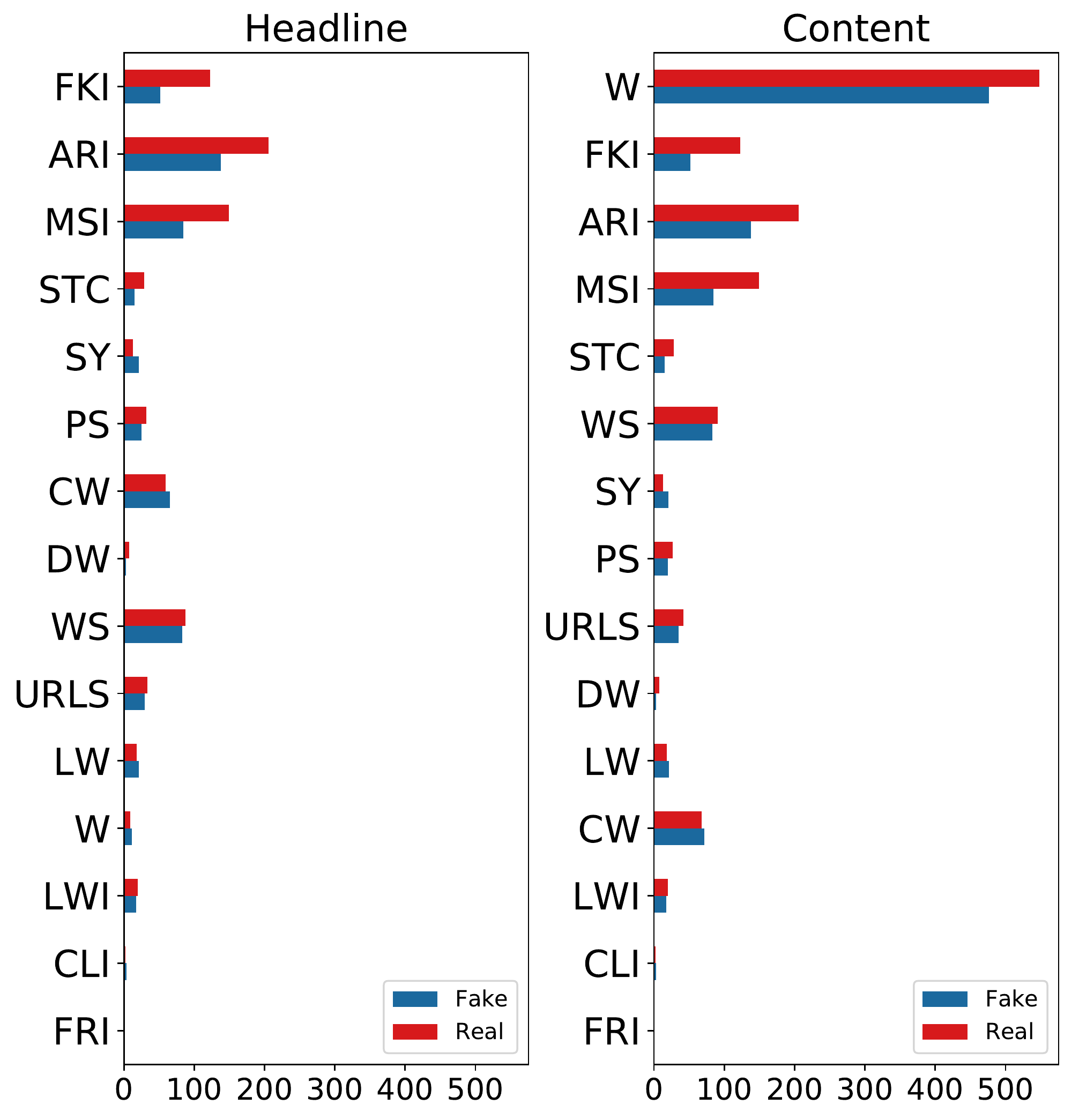}
	}
	\subfigure[Web-markup]{
		\label{fig:top15_dist_web}
		\includegraphics[width=20mm,height=45.4mm]{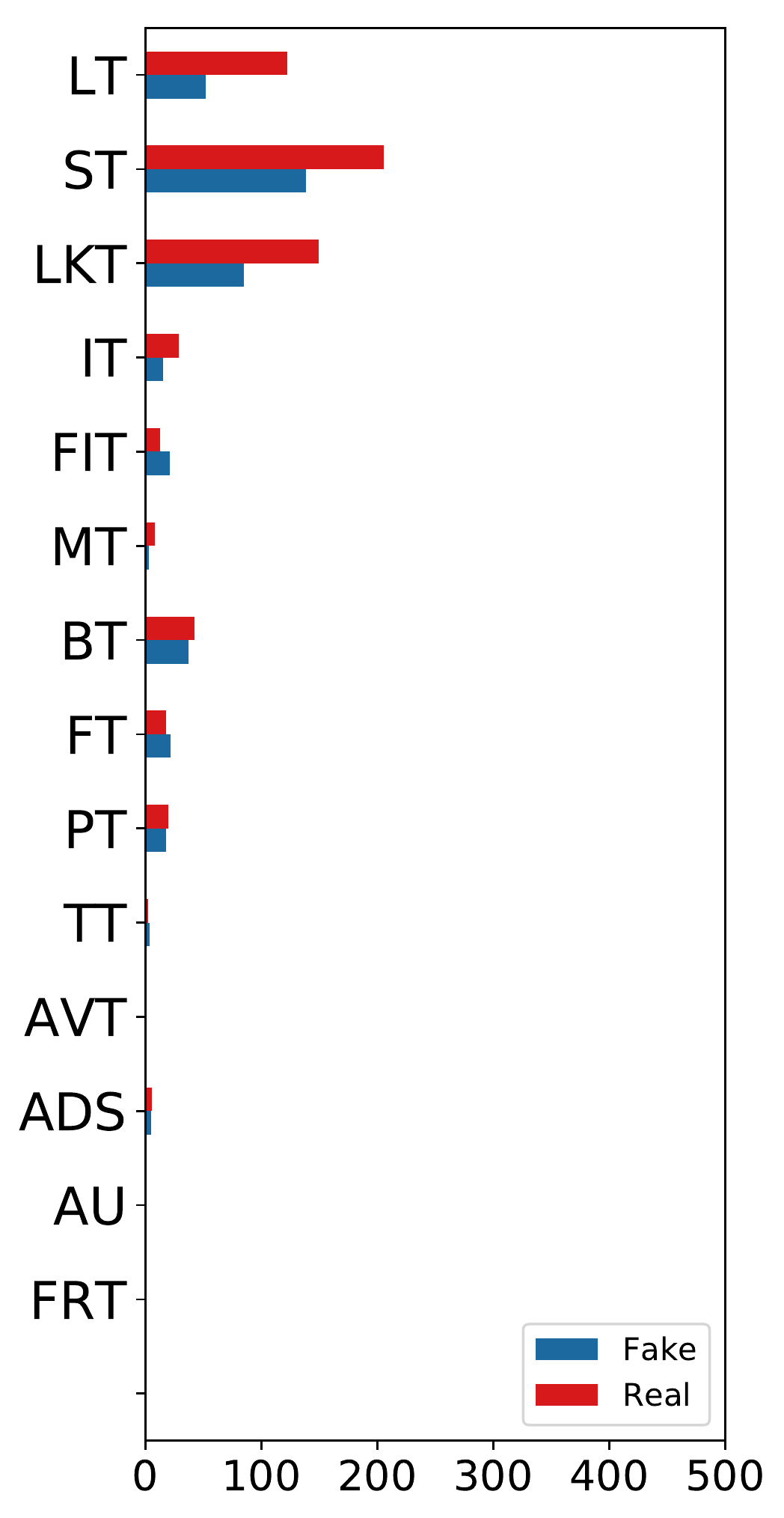}
	}
	\vspace{-11pt}	
	\caption{Mean frequency distribution of features per class in the \ourdata \dataset.}
	\vspace{-6pt}
	\label{fig:features}
\end{figure*}

\subsection{Topic-Agnostic Features}
\label{sec:features}
While exploring web pages containing fake news, we observed some salient topic-agnostic features. For example, the pages contain a large number of ads -- this is not surprising given that providers can gain significant advertising revenue by attracting users to their web site with appealing fake news headlines ~\citep{allcott2017social}.
Recent works have also argued that fake news articles are designed to induce inflammatory emotions in readers, and contain text patterns related to understandability that differ from mainstream news~\cite{horne2017just, perez@coling2018, bakir2018fake}.

Figure~\ref{fig:fn_screenshots} shows some examples of fake and real news pages. Fake news pages not only have a larger number of ads and polluted layouts but also have a distinctive style to their headlines, often in the form of a sensationalist slant. Additionally, besides attempting to describe the article, these headlines often contain terms designed to catch the readers' attention (e.g., ``Just in'', ``Read this'', ``Breaking News'').
These observations motivated us to investigate two broad classes of features: web-markup and linguistic-based (morphological, psychological and readability-related). The features are listed in Table~\ref{table:topic_agnostic_features} and we summarize them below.

\paragraph{Morphological Features.}
This set corresponds to the frequency (word count) of morphological
patterns in texts. We obtain these patterns through part-of-speech tagging, which
assigns each word in a document to a category based on both its
definition and context. 

\paragraph{Psychological Features.}
Psychological features capture the percentage of total semantic words in
texts\footnote{http://liwc.wpengine.com/interpreting-liwc-output/}. We obtain the words' semantics by using a dictionary that has
lists of words that express psychological processes (personal
concerns, affection,
perception). 

\paragraph{Readability Features.}
This set captures the ease or difficulty of comprehending a text.
We obtain these features through readability scores and counting of character, words, and sentences usage.

Previous works have found that fake news often displays a divergence
between the news headline and the body
content~\cite{horne2017just,silverman2015lies}: (i) a headline declares a piece of information to be
false and the body text declares it to be true (or vice-versa); and
(ii) fake news packs the main claim of the article into its title,
allowing the reader to skip reading the body article, which tends to
be short, repetitive, and less informative when compared with real
news.
These divergences between the textual pieces of news articles
motivated us to apply linguistic features at different granularities:
considering only the headline, only the content, and combining
both. 
Figures~\ref{fig:top15_dist_nltk}, ~\ref{fig:top15_dist_liwc}, and ~\ref{fig:top15_dist_readby}
show the ranking of the top-15 linguistic-based features, in \ourdata \dataset, with the
largest differences between classes considering distinct text
granularities.
In Figure~\ref{fig:features}, the psychological features are named
alongside an index. This is because each of the psychological features
represents a category that contains other features. A complete list of
these features is available at \url{http://liwc.wpengine.com/compare-dictionaries}.

\paragraph{Web-Markup Features}. 
These features capture patterns of the web pages' layout . The web-markup features we use include: frequency (number of occurrences) of advertisements,
presence of an author name (binary value), and the frequency of distinct categories of tag groups\footnote{https://www.w3schools.com/tags/ref\_byfunc.asp}.
Figure~\ref{fig:top15_dist_web} shows the mean distribution frequency of each web-markup features in \ourdata \dataset, within fake and real news.
Note that the distributions are different, thus supporting the use of these features to distinguish news.

\subsection{Feature Selection}
We perform feature selection using a combination of four different methods: Shannon
Entropy~\cite{lesne2014shannon}, Tree Based Rule~\cite{geurts2006extremely}, L1 Regularization~\cite{zhao2006model} and Mutual Information~\cite{kraskov2004estimating}. We combine the outputs from these methods by normalizing them and applying the geometric mean to obtain a new score $r(f_{i})$ which corresponds to the importance of a feature $f_{i}$:
\begin{equation}
r(f_{i})  = \sqrt[4]{\rule{0pt}{0.9em} SE(f_{i})^{-1} \times TB(f_{i}) \times L1(f_{i}) \times MI(f_{i})}
\end{equation} 
where SE refers to Shannon Entropy, TB to Tree Based Rule, L1 to L1
Regularization and MI to Mutual Information. We compute the
feature importance scores and remove features with $r(f_{i})$ value equal to
$0$. When we performed this process for features from news headlines,
we found the following patterns turned out to be ineffective: FOW, IN, JJR, PRP\$, TO, VBD,
VBG, VBZ, WP\$, MSI, CW, TT, FW (semicolon), BP (ingest), RL (time)
and PC (home). When we consider features from news content, DT, PDT, RBR, RP,
OG, and UH are removed. When we extract features from both
(headline + content), DT, JJS, PDT, POS, RBR, RBS, UH and WRB are
eliminated.
For the web-markup features, we kept IT, AVT, AU, LKT, ADS, ST and BT. 
The sizes of the sets of  topic-agnostic features are:
(1) for headlines, 137 features; (2) for content, 148; and (3) headline+content, 145. 
We study the effectiveness of these features in Section~\ref{sec:exp}.
Note that we performed the feature selection process only once, using the 
\ourdata \dataset, and used the resulting features on all experiments.

\subsection{Classification}
We use supervised learning to classify news based on the selected features. Formally, the resulting classifier corresponds to a function that takes as input the TAG features of a web page $x$ $\epsilon $ $\mathbb{R}^{d}$ and produces an output $\hat{y}$ $\epsilon$ $C$, where $C$ is the set of all categories. Here we use two categories: fake and real news. 
We experimented with three different learning methods: Support Vector Machine (SVM), K-Nearest Neighbors (KNN) and Random Forest (RF). We use SVM classifier with linear kernel and cost equal to \textit{$0.1$}. The other methods are setup with default parameters of the scikit-learn\footnote{http://scikit-learn.org/} library.

\section{Experimental Evaluation}
\label{sec:exp}
In this section, we describe the evaluation that we applied to assess how effective our approach is at identifying fake news.

\paragraph{Corpus.} 
As discussed in Section~\ref{sec:related}, existing public \datasets are limited. So, we built a new dataset that we call \textbf{\emph{\ourdata}}\footnote{https://osf.io/ez5q4/}, by compiling a list of known reliable and unreliable sites and crawling them.
As seeds for the crawl, we used web sites coming from Politifact\footnote{https://www.politifact.com/punditfact/article/2017/apr/20/politifacts-guide-fake-news-websites-and-what-they/}, BuzzFeed~\cite{buzzfeed2017git}, and OpenSources.co\footnote{http://www.opensources.co} as unreliable news sites and 242 sites from the most visited news sites {\it Alexa's top 500 news sites}\footnote{https://www.alexa.com/topsites/category/News} as reliable sites. Since the last one is based just on popularity, we manually selected the web sites focused on political news.  One challenge in constructing such a large dataset is the impracticality of individually labeling all the articles. Our approach was to project the domain-level ground truth onto the content collected from those domains.
We collected over 1.6 million web pages published between 2013 and 2018. Then, we post-processed the data removing non-political pages using a Na\"{i}ve Bayes model combined with TF-IDF feature representation, and as a training set, we use a publicly-available  corpus\footnote{https://www.kaggle.com/rmisra/news-category-dataset/home}. Furthermore, to ensure a balanced distribution of web pages sites for each year, we sampled 32 pages from each site for each of the years we have crawl data for.
The result of this balancing is a \dataset of 14,240 news pages with 7,136 pages coming from 79 unreliable sites, and another 7,104 coming from 58 reliable sites. The motivation for this balancing is to avoid the problem of overfitting (since real news is orders of magnitude more prevalent than fake news). Additionally, ensuring balance across the years covered allows us to evaluate the effectiveness of different approaches as news topics change over time.

To demonstrate the robustness of the TAG model, we also evaluate our approach over two additional \datasets that have been used in previous works:  \textbf{\emph{Celebrity}}~\cite{perez@coling2018} and  \textbf{\emph{\uselectiondata}}~\cite{allcott2017social}. We note that these \datasets contain only the text of the web pages. Since we need to extract web-markup features, we fetched the original web page source from the Web. For sites and pages that were no longer available, we retrieved versions from the Web Archive. In addition, duplicated articles were removed. As a result, the versions of the Celebrity and \uselectiondata \datasets\footnote{https://osf.io/qj86g/} used in our experiments contain 479 and 691 news articles, respectively.

\paragraph{Experimental Setup.}
We use the NLTK \cite{bird2004nltk} library part-of-speech tagger to compute  \emph{Morphological Features}. To extract the  \emph{Psychological Features}, we use the Linguistic Inquiry and Word Count software (LIWC, Version 1.3.1 2015) \cite{pennebaker2015liwc}. To compute \emph{Readability Features}, we use a Textstat\footnote{http://pypi.python.org/pypi/textstat/} library, and finally, to extract the \emph{Web-Markup Features} from web pages, we use BeautifulSoup\footnote{https://www.crummy.com/software/BeautifulSoup} and Newspaper.\footnote{https://newspaper.readthedocs.io/en/latest}

We compare the TAG classifier with the \michiganapproach (\michiganapproachabbrev) presented in \cite{perez@coling2018}. This model represents documents by using four sets of linguistic
features: n-grams (unigrams + bigrams encoded by TF-IDF values), psychological, readability and syntactical features (production rules of context-free grammars encoded by TF-IDF values). 
The psychological and readability sets are the same as we use (see Table \ref{table:topic_agnostic_features}).
We selected the \michiganapproachabbrev for two main reasons: like our approach, they focus on the automatic identification of fake content in online news; and because their classifier attained high accuracy using content-based features, it is a suitable baseline for our TAG classifier. To better understand the contribution of individual features, we created multiple baseline classifiers using different feature combinations. We used a linear SVM classifier and conducted our evaluations using five-fold cross-validation with accuracy as the performance metric.

\subsection{Effectiveness of Different Features}
\label{features_exp}

\setlength{\tabcolsep}{3pt}
\begin{table}[!htb]
	\centering
	\small
	\caption{Accuracy results for models  that use different set of topic-agnostic features -- where L is LIWC, N is NLTK, R is readability, and W is webmarkup features -- over three different \datasets: Celebrity, \uselectiondata, and \ourdata. The best accuracies for each feature set are \textbf{bold}; the best accuracies for each news article's representations (H, C and HC) are \underline{underlined}.}
	\vspace{-6pt}
	\begin{tabular}{ l  c c c | c c c  | c c c} 
		\toprule[0.4mm]
		\textbf{\Dataset}&\multicolumn{3}{c}{\textbf{Celebrity}} &\multicolumn{3}{c}{\textbf{\uselectiondata}} &\multicolumn{3}{c}{\textbf{\ourdata}}\\
		\textbf{Features}&\textbf{H} &\textbf{C}&\textbf{HC} &\textbf{H} &\textbf{C}&\textbf{HC} &\textbf{H} &\textbf{C}&\textbf{HC}\\
		\midrule[0.4mm]
		\textbf{W}& \textbf{0.68}&\textbf{0.68}&\textbf{0.68} & \textbf{0.65}&\textbf{0.65}&\textbf{0.65} & \textbf{0.71}&      \textbf{0.71}& \textbf{0.71}  \\
		\textbf{L} & 0.69&\textbf{0.73}&\textbf{0.73} & 0.77&0.81&\textbf{0.83} &        0.71&        0.75& \textbf{0.76}\\
		\textbf{N} & 0.58&\textbf{0.68}&0.66 & \textbf{0.81}&0.75&0.76& \textbf{0.77}&            0.66&      0.67 \\
		\textbf{R} & 0.57&\textbf{0.62}&0.57 & \textbf{0.75}&0.73&0.73 & \textbf{0.69}&      0.62&      0.64\\
		
		\midrule
		\textbf{L-R} & 0.65&\textbf{0.76}&0.74 & 0.79            &0.82&\textbf{0.83} & 0.74              &0.75&\textbf{0.76}\\
		\textbf{N-R}& 0.65&\textbf{0.68}&0.67 &\textbf{\underline{0.83}}&0.78&0.78              & \textbf{0.78}  &0.71             &0.72 \\
		\midrule
		\textbf{N-W} & 0.68&\textbf{0.72}&\textbf{0.72} & 0.79&0.79&\textbf{0.80} & \textbf{0.81}&      0.79&      0.79 \\
		\textbf{L-W} & 0.70&\textbf{0.77}&\underline{0.75} & 0.79&0.82&\textbf{0.83} & 0.78&      \textbf{\underline{0.81}}&      \textbf{0.81} \\
		\textbf{R-W} & 0.67&\textbf{0.72}&0.67 & \textbf{0.80}&0.76&0.76 & \textbf{0.77}&     0.76 &0.76 \\
		\textbf{N-R-W} & 0.67&\textbf{0.72}&0.71 &\textbf{\underline{0.83}}&0.79&0.80 &\textbf{0.81}&0.78&0.79\\
		\textbf{L-R-W} &0.71&\textbf{\underline{0.78}}&0.71 & 0.79&\underline{0.83}&\textbf{0.85} & 0.80&0.80&\textbf{0.81} \\
		\textbf{L-N-R-W}& \underline{\textbf{0.73}}&\textbf{0.73}&0.71  & \underline{0.83}&0.82&\textbf{\underline{0.86}} & \textbf{\underline{0.83}}&\underline{0.81}&\underline{0.82}\\
		\bottomrule[0.4mm]
	\end{tabular}
	\label{table:table_compare_features_all}
	
	\vspace{10pt}
	
	\centering 
	\small
	\caption{Classification results (accuracies) for three \datasets.}	
	\vspace{-6pt}
	\begin{tabular}{ l  c c c c}
		\toprule[0.4mm]
		\Dataset & \textbf{Celebrity} & \textbf{\uselectiondata} & \textbf{\ourdata} \\
		\midrule
		\textbf{\michiganapproachabbrev} &0.73 & 0.81 & 0.76\\
		\textbf{TAG Model}&\textbf{0.78}&\textbf{0.86}  & \textbf{0.83}\\
		\bottomrule[0.4mm]
	\end{tabular}
	\vspace{-10pt}
	\label{table:table_compare_features_baseline}
\end{table}

We evaluate the performance of models trained with different combinations of feature sets (separately and jointly). 
In addition, to assess the performance of the classifiers using different representations for a news article, each experiment was performed for the headline (H),  content (C) and the combination of both (HC).

Table~\ref{table:table_compare_features_all} shows the accuracy obtained for the different TAG feature sets over the three \datasets. Note that combining features often leads to the highest accuracies.  For example, for political news, the highest accuracies are obtained by the configuration that combines \textbf{LIWC (L) - NLTK (N) - readability (R) - webmarkup (W)}, which attains 0.86 accuracy for \uselectiondata and 0.83 for \ourdata. For the Celebrity data, the best configuration is \textbf{LIWC (L) - readability (R) - webmarkup (W)} which attains 0.78 accuracy.
If we further examine the results from LIWC, we can see that combining LIWC with other features often leads to higher accuracies: for all \datasets, the accuracy gains vary between 0.012 and 0.21. This reinforces previous findings~\cite{perez@coling2018, bakir2018fake} that psychological factors play an important role in the disinformation ecosystem. The web-markup features also lead to improvements, in particular when compared against readability for the \ourdata corpus.

When we consider the features from different article parts (H, C, and HC), the distribution of accuracies for \uselectiondata and \ourdata are similar. This suggests that the headlines contain useful information that allows the identification of political fake news. Thus, either headlines or content or both can be used for classification.
On the other hand, for the Celebrity \dataset, we can clearly see that the classifier achieves slightly better results using the content of the articles. Consider the following two examples from this dataset: (1) a real news, where the headline is ``\emph{Stop Right Now! The Spice Girls Might Be Planning A Reunion}'' and the content is ``\emph{There are certain things that people just shouldn't joke about. A potential Spice Girls reunion with all five members is one of them. Earlier this morning...}'', and (2) a fake news, where the headline is ``\emph{Taylor Swift Goes Naked in `...Ready for It?' Watch!}'' and the content is ``\emph{Nope, definitely not ready for this! Taylor Swift released a 15-second teaser for the music video for her new song `... Ready for It?' on Monday...}''. 
The previous example in the Celebrity dataset shows a concrete example where headlines (H)
for fake and real news have similar linguistic features (e.g., use of capitalized words, punctuation), but content (C) is more discriminative in terms of linguistic features and readability. Furthermore, the web-markup features are effective in this scenario, since they make it possible to detect a clear difference between fake and real news based on web page characteristics (e.g., number of ads, links, and images).

We also compare the best TAG feature set combination for each dataset, identified during our previous experiment, with the baseline. We use a linear SVM and five-fold cross-validation with accuracy as metric. 
As shown in Table~\ref{table:table_compare_features_baseline}, the classifier based on TAG features outperforms the \emph{\michiganapproachabbrev} for all \datasets,  indicating that the task of fake news classification can be effectively accomplished using topic-agnostic features. Furthermore, it is  important to note that \emph{\michiganapproachabbrev} features uses 3 orders of magnitude more features (\textbf{$\sim$800,000}) than our approach (\textbf{$\sim$160} features). The number of features has important implications for the processing time, maintaining the model, and explainability.

\subsection{Effectiveness over Time}
\label{sec:time_exp}

To study the behavior of our approach as news topics change over time, 
we split \ourdata \dataset into six time windows (sub-\datasets) corresponding to each year from 2013 to 2018.
We then experimented with multiple classifier configurations: each configuration uses one sub-\dataset for training and the others (one at a time) for testing. 
For example, we use pages published in 2013 to construct a classifier $C_{2013}$ and use $C_{2013}$
to classify pages in the sub-datasets from 2014 through 2018.
Note that each sub-\dataset is associated with 5 results. For this experiment, our TAG model uses the set of features NLTK, LIWC, readability and web-markup.

\begin{figure}[t]
	\centering
	\includegraphics[width=0.70\columnwidth]{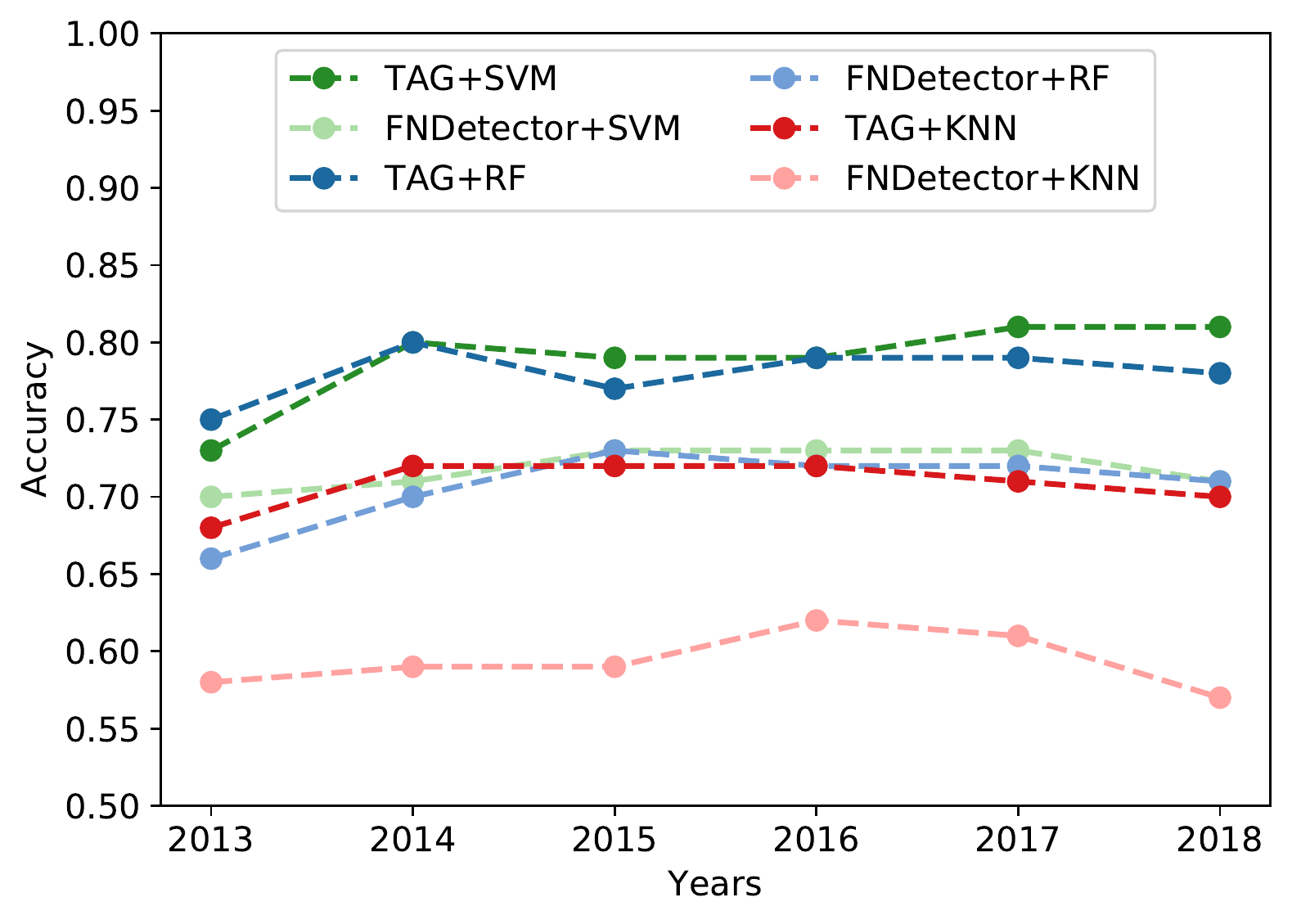}
	\vspace{-.5cm}						
	\caption{Effectiveness of SVM, KNN and RF using baseline (\michiganapproachabbrev) and TAG model in different time windows.}
	\label{fig:results_perfomance_by_time}
	\vspace{-.4cm}						
\end{figure}

\begin{figure}[t]
	\centering
	\subfigure[2013]{\label{fig:a}\includegraphics[width=27.5mm]{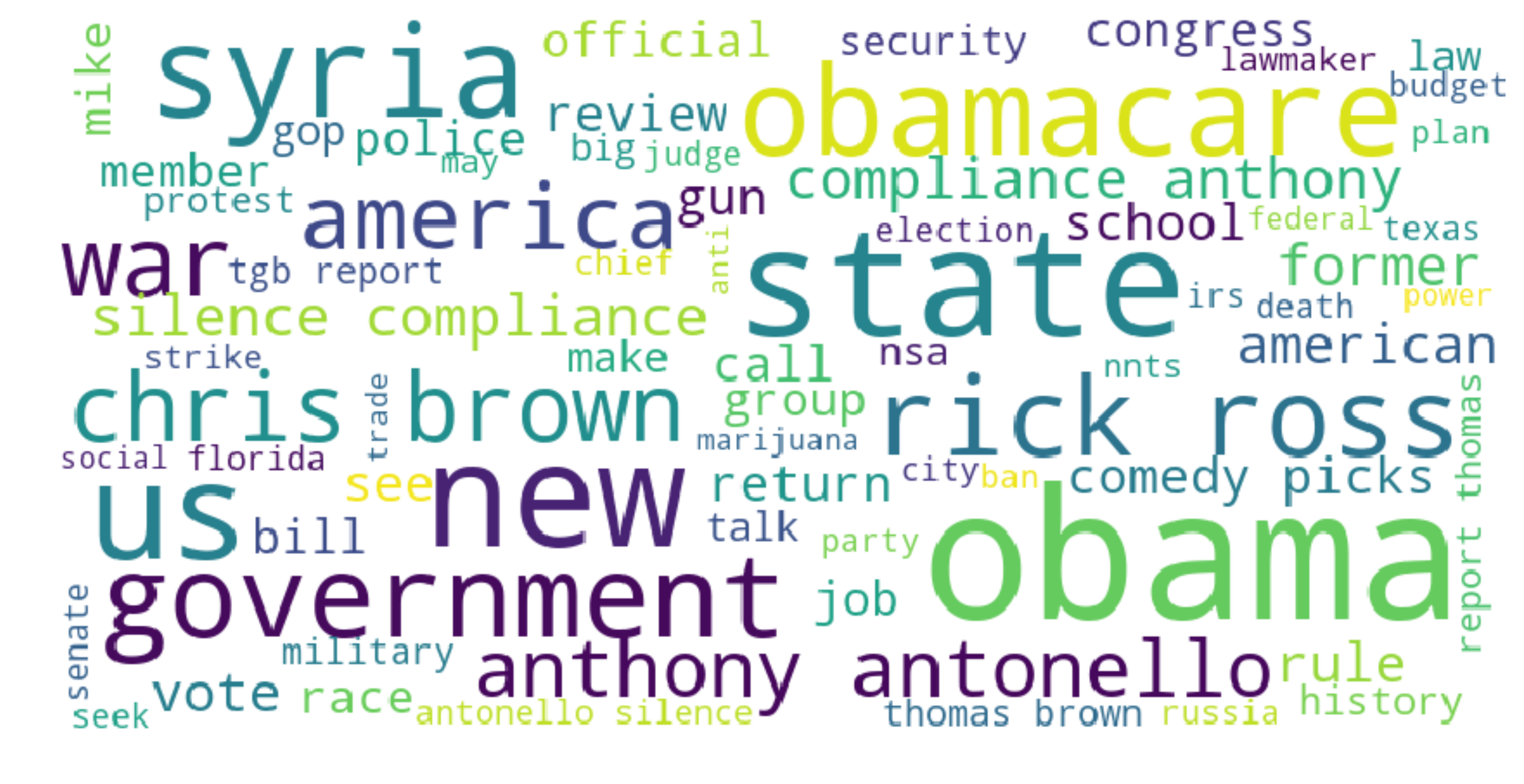}}
	\subfigure[2014]{\label{fig:b}\includegraphics[width=27.5mm]{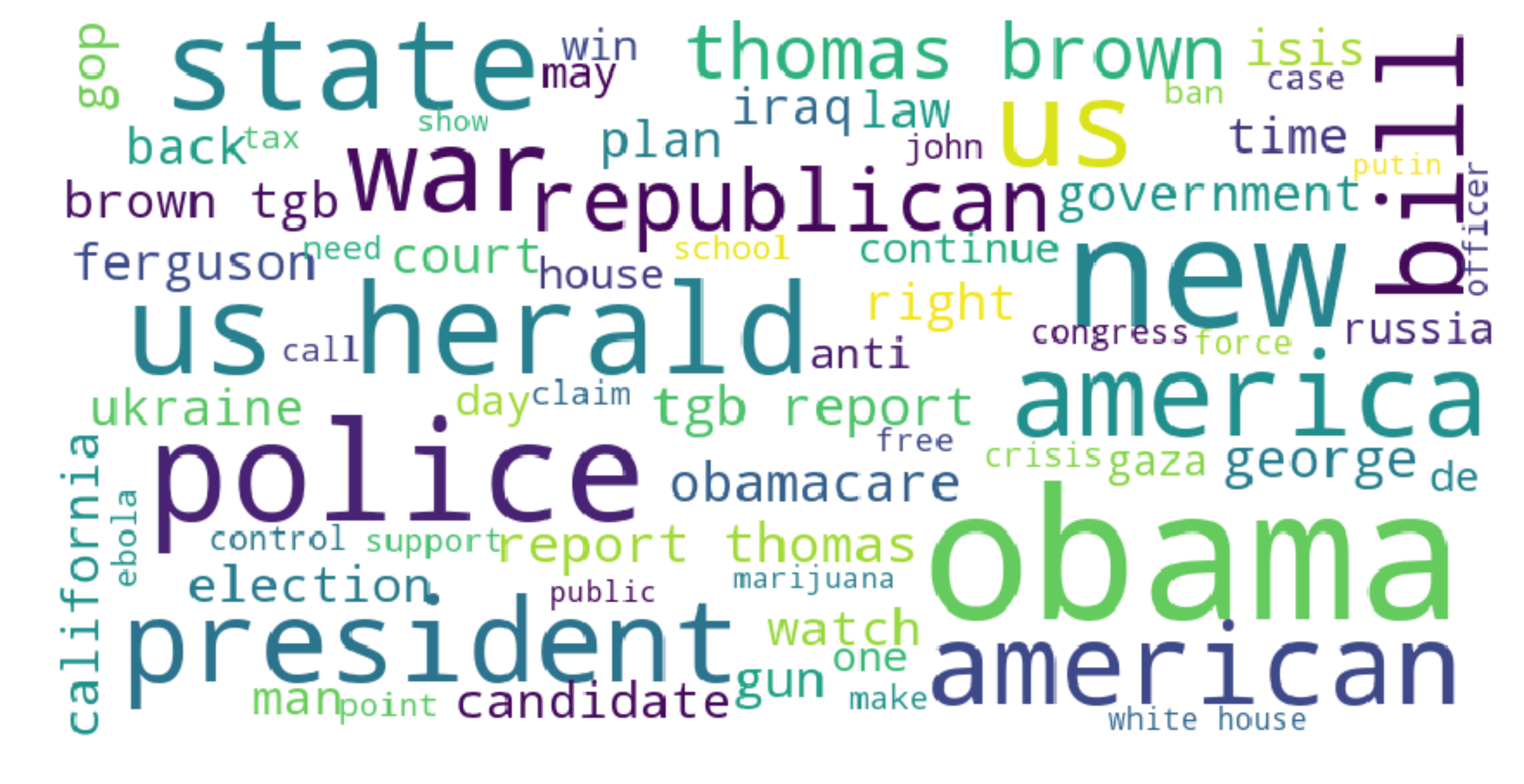}}
	\subfigure[2015]{\label{fig:d}\includegraphics[width=27.5mm]{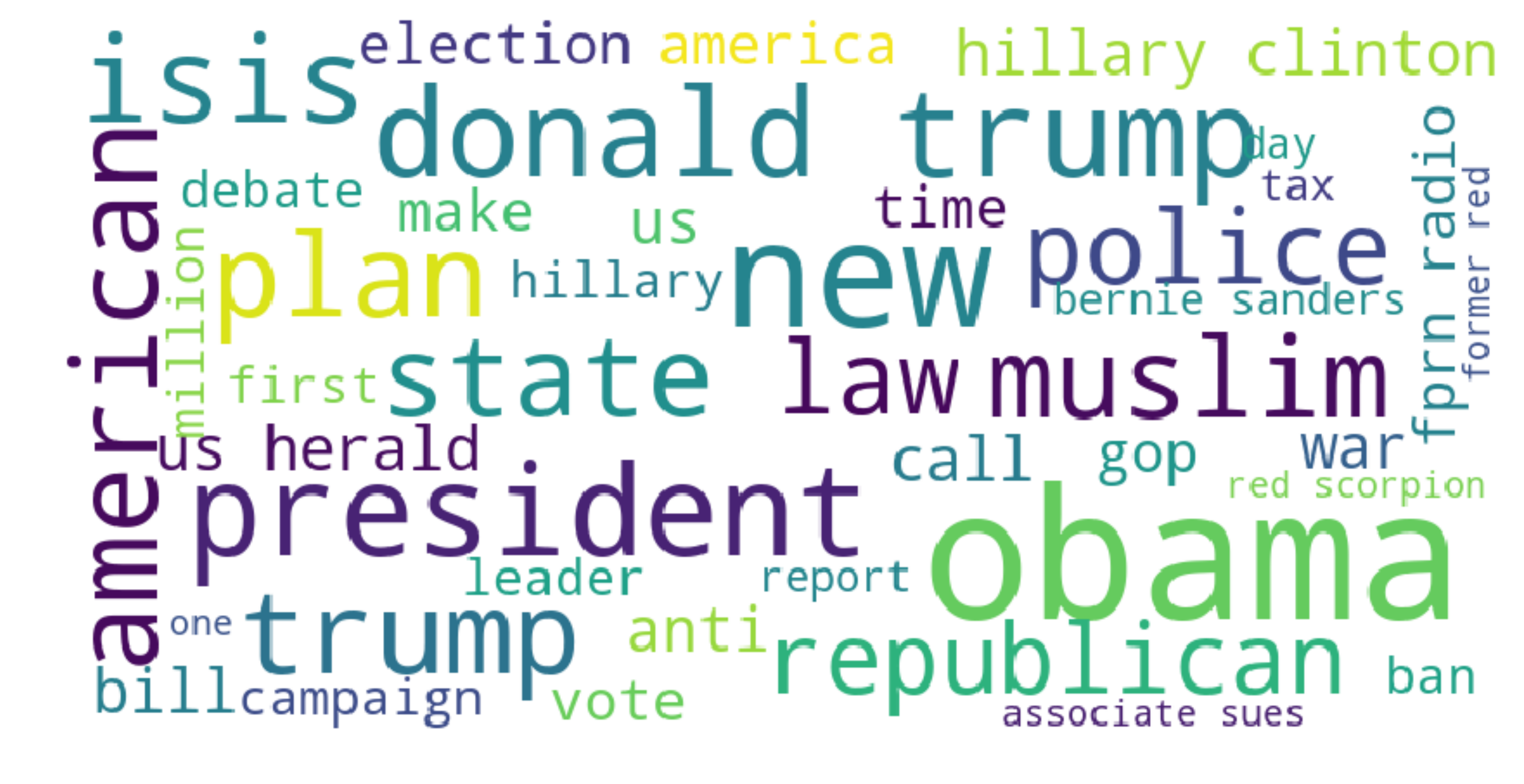}}
	\vspace{-.15cm}		
	\subfigure[2016]{\label{fig:e}\includegraphics[width=27.5mm]{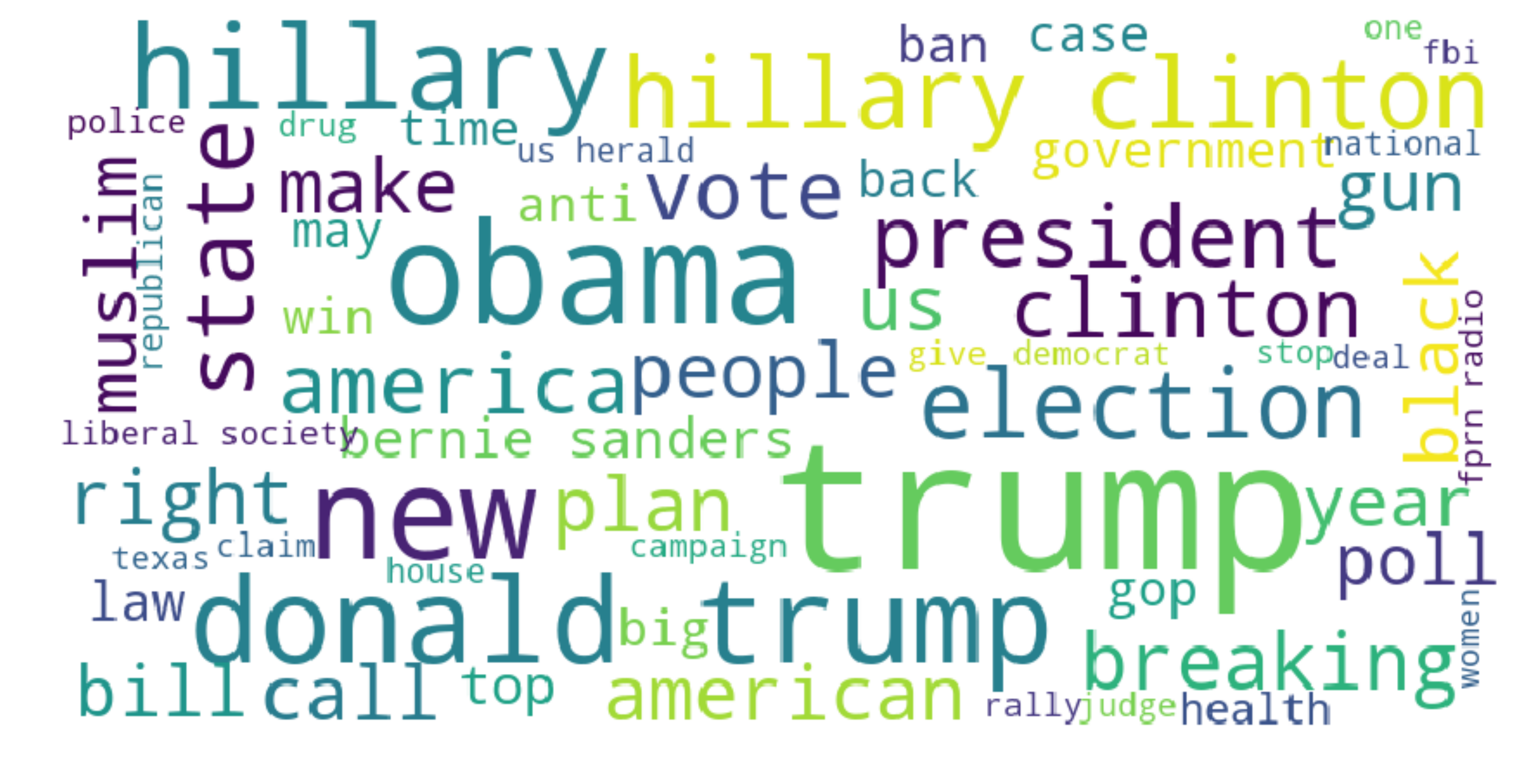}}
	\subfigure[2017]{\label{fig:f}\includegraphics[width=27.5mm]{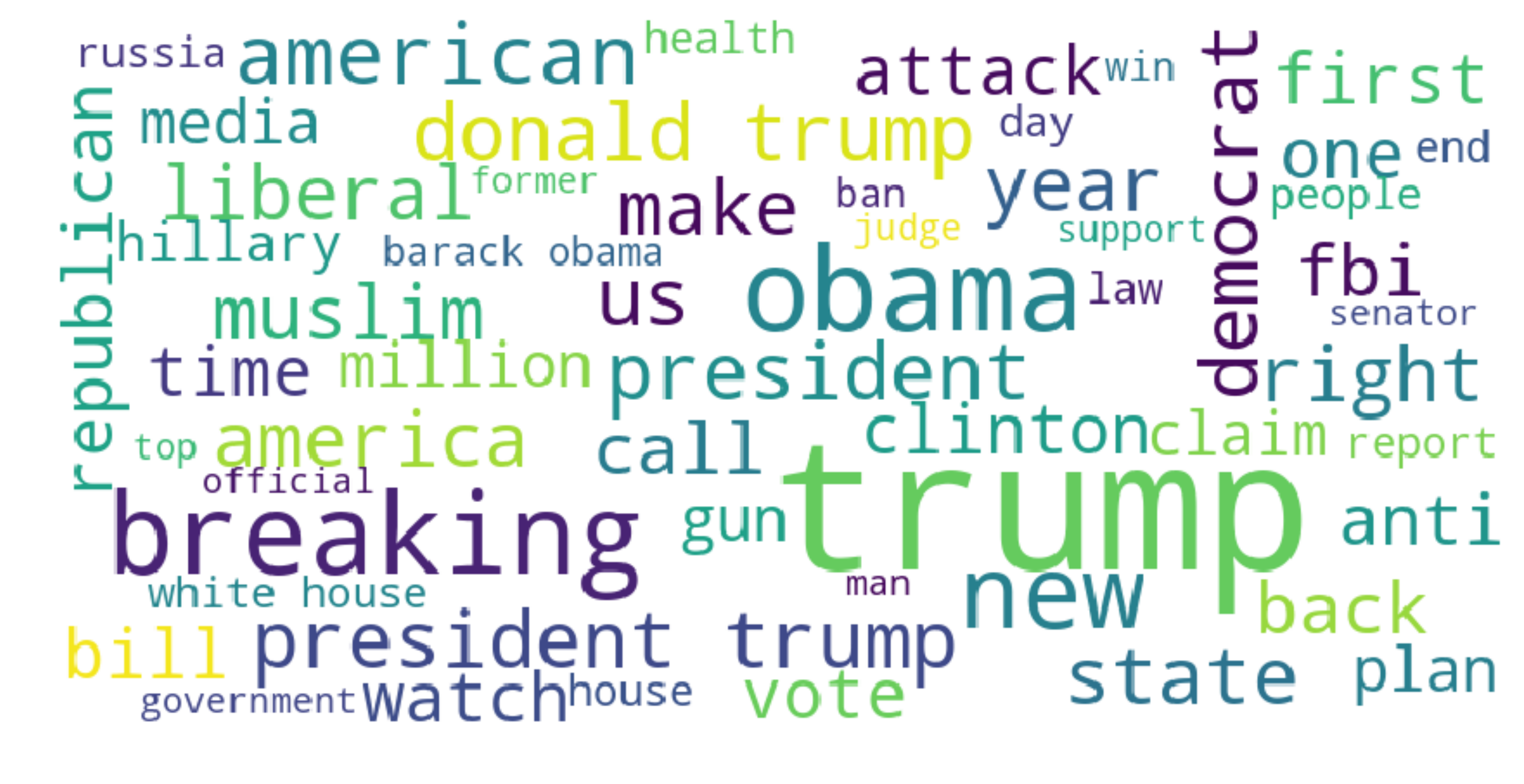}}
	\subfigure[2018]{\label{fig:g}\includegraphics[width=27.5mm]{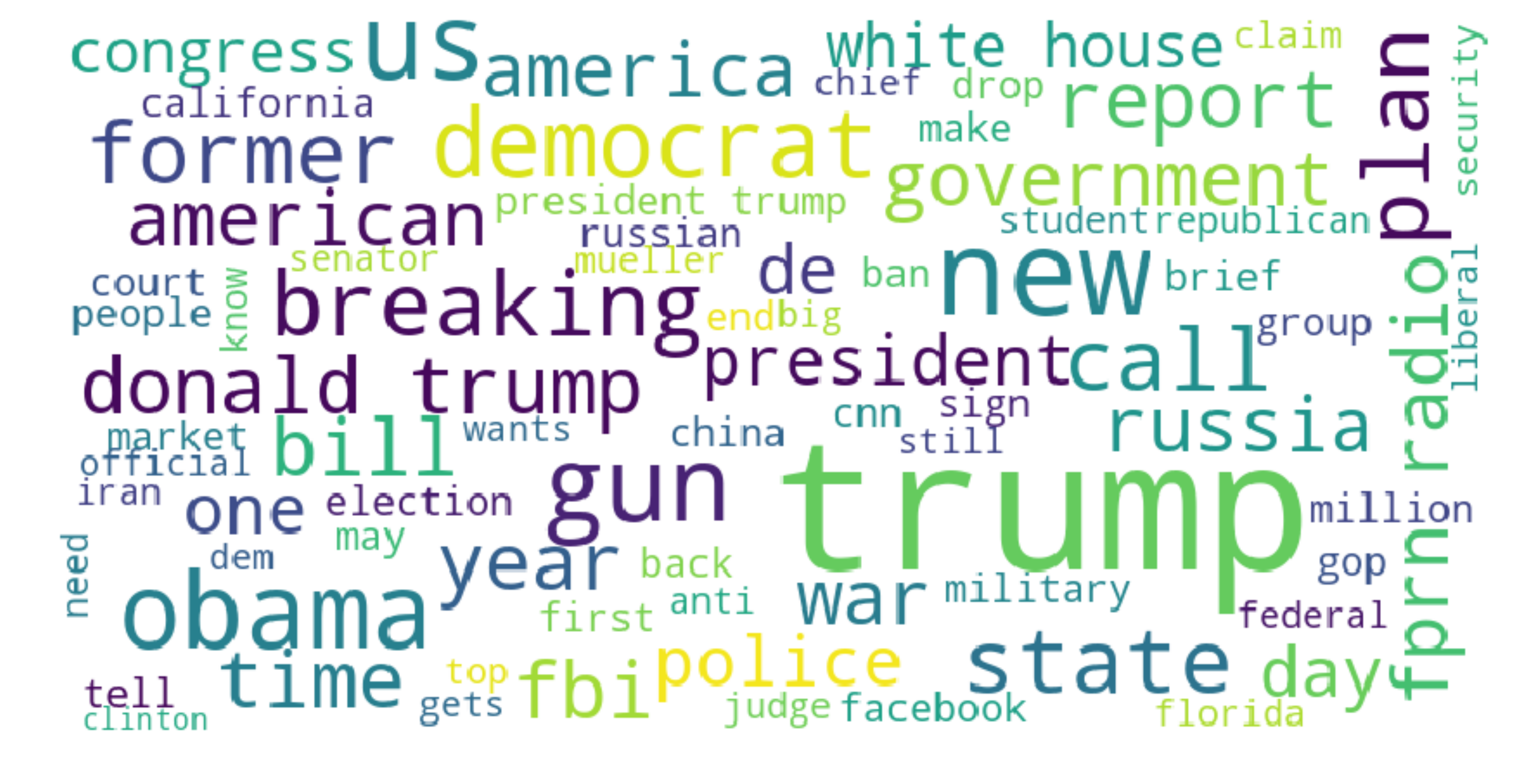}}	
	\vspace{-.4cm}						
	\caption{Tag clouds of the most frequent terms in web pages.}
	\label{fig:word_cloud_per_year}
	\vspace{-.2cm}						
	
\end{figure}

Figure~\ref{fig:results_perfomance_by_time} shows the average accuracies for each sub-\dataset over the 5 test sets. 
Our TAG model performs better than \emph{\michiganapproachabbrev} for all the time windows. The relatively lower accuracy values obtained by \emph{\michiganapproachabbrev} can be explained by its dependence on the contents.
You can observe the topic changes in Figure~\ref{fig:word_cloud_per_year}, which shows tag clouds (built over our training data using n-grams frequencies) summarizing the news in each year. Note that even though the tag clouds share some keywords (e.g., ``Obama'' and ``Trump''), they appear at different frequencies and cover distinct subjects. 

The content-based model essentially learns how to detect fake news for a specific time and topics. Note, for example, the difference between the topics in 2013 and in 2015. In 2013, ``Trump'' was not yet involved in politics (see Figure~\ref{fig:word_cloud_per_year}(a)), but he starts to appear in political news in 2015, and more frequently since the 2016 elections
as shown in Figure~\ref{fig:word_cloud_per_year}(e) and Figure~\ref{fig:word_cloud_per_year}(f). 
Also note the differences between 2016 and 2017: 
In 2016, political news were centered around the campaign, parties, voters, etc., and we see terms such as ``Hillary'', ``Clinton'', ``vote'', ``Trump'' and ``Obama''. But in 2017 (after the elections), the terms ``Hillary'', ``Clinton'' decrease and terms like ``breaking'', ``attack'', ``Russia'', ``FBI'' start to appear in the political discourse.
This indicates that, to be effective, content-based models must be constantly retrained. This is both costly and error-prone.

\subsection{Effectiveness for Different Domains} 
\label{sec:topic_exp}
We also evaluated the generalizability of our approach when the training and testing sets are drawn from entirely different topic domains.
We ran two experiments: in the first, we use Celebrity as a training \dataset and \uselectiondata as testing \dataset;  and for the second experiment, the other way around. We use all the TAG feature sets identified during our previous experiments (N, L, R and W). 
We considered three classifiers -- SVM, KNN and RF; and used five-fold cross-validation.
The results in Table~\ref{table:domain_agnostic} show that our topic-agnostic approach attains 
accuracies that are substantially higher than those of \michiganapproachabbrev. 
While our original goal in this work was to design a classifier that is able to distinguish fake and real political news as they evolve over time, 
these results show that the approach is promising for topic domains beyond political news.

\section{Conclusions \& Future Work}
\label{sec:conclusion}
We presented a new approach to detect fake news web pages which uses topic-agnostic features.
Through a detailed experimental evaluation, we showed that our approach accurately classifies not only political news as topics evolves over time but also news from different domains, outperforming content-based approaches while using significantly fewer features and requiring no frequent re-training. Our results suggest that topic-agnostic features are effective for distinguishing between fake and real news, and that robust classifiers can be constructed that enable 
the timely discovery of fake news articles.
We have also created a new corpus of over 14,000 political news pages drawn from 137 sites and spanning 6 years. To the best of our knowledge, this is the first of its kind in terms of size, focus on the Web, and inclusion of HTML markup information. 

There are many directions we plan to pursue in future work. We will explore further improvements to the topic-agnostic model by considering additional features, for example, user engagement and network structure. We would also like to experiment with different strategies to expand our fake news corpus, including the use of social media to search for previously unknown sites, and by using the top-agnostic classifier in conjunction with a focused crawler to discover new sites as they are created.

\begin{table}[t]
	\small
	\caption{Cross-domain results (accuracies) between models.}	
	\vspace{-8pt}
	\begin{tabular}{cclcc}
		\toprule[0.3mm]
		\multirow{2}{*}{\textbf{Training}}   & \multirow{2}{*}{\textbf{Test}}  & \multirow{2}{*}{\textbf{Classifier}} & \multicolumn{2}{c}{\textbf{Accuracy} }\\ 
		& &  & \begin{tabular}[c]{@{}l@{}}\textbf{} \textbf{\michiganapproachabbrev}\end{tabular} & \begin{tabular}[c]{@{}l@{}}\textbf{}\textbf{TAG Model}\end{tabular} \\ 
	\midrule[0.3mm]
		\multirow{3}{*}{Celebrity} &\multirow{3}{*}{\begin{tabular}[c]{@{}l@{}}US-Election\\ 2016\end{tabular}} & SVM & 0.59 & \textbf{0.70} \\ 
		& & KNN & 0.59 & \textbf{0.64}       \\ 
		& & RF    &  0.56 & \textbf{0.64}       \\ \hline
		\multirow{3}{*}{\begin{tabular}[c]{@{}l@{}}US-Election\\ 2016\end{tabular}} & \multirow{3}{*}{Celebrity} & SVM &0.59 &  \textbf{0.63}  \\ 
		& & KNN&0.56&\textbf{0.60 } \\ 
		& & RF &0.51 & \textbf{0.60}   \\ 
		\bottomrule[0.3mm]
	\end{tabular}
	\vspace{-10pt}
	\label{table:domain_agnostic}
\end{table}

\begin{acks}
	This work was partially supported by the DARPA MEMEX and D3M programs, NSF award CNS-1229185, and the Brazilian National Council for the Improvement of Higher Education (CAPES) under grant 88887.130294/2017-00. Any opinions, findings, and conclusions or recommendations expressed in this material are those of the authors and do not necessarily reflect the views of the sponsors.
\end{acks}

\bibliographystyle{ACM-Reference-Format}
\balance 
\bibliography{paper}


\begin{thebibliography}{23}


\ifx \showCODEN    \undefined \def \showCODEN     #1{\unskip}     \fi
\ifx \showDOI      \undefined \def \showDOI       #1{#1}\fi
\ifx \showISBNx    \undefined \def \showISBNx     #1{\unskip}     \fi
\ifx \showISBNxiii \undefined \def \showISBNxiii  #1{\unskip}     \fi
\ifx \showISSN     \undefined \def \showISSN      #1{\unskip}     \fi
\ifx \showLCCN     \undefined \def \showLCCN      #1{\unskip}     \fi
\ifx \shownote     \undefined \def \shownote      #1{#1}          \fi
\ifx \showarticletitle \undefined \def \showarticletitle #1{#1}   \fi
\ifx \showURL      \undefined \def \showURL       {\relax}        \fi
\providecommand\bibfield[2]{#2}
\providecommand\bibinfo[2]{#2}
\providecommand\natexlab[1]{#1}
\providecommand\showeprint[2][]{arXiv:#2}

\bibitem[\protect\citeauthoryear{Allcott and Gentzkow}{Allcott and
  Gentzkow}{2017}]%
        {allcott2017social}
\bibfield{author}{\bibinfo{person}{Hunt Allcott} {and} \bibinfo{person}{Matthew
  Gentzkow}.} \bibinfo{year}{2017}\natexlab{}.
\newblock \showarticletitle{Social media and fake news in the 2016 election}.
\newblock \bibinfo{journal}{\emph{Journal of Economic Perspectives}}
  \bibinfo{volume}{31} (\bibinfo{year}{2017}), \bibinfo{pages}{211--36}.
\newblock


\bibitem[\protect\citeauthoryear{Bakir and McStay}{Bakir and McStay}{2018}]%
        {bakir2018fake}
\bibfield{author}{\bibinfo{person}{Vian Bakir} {and} \bibinfo{person}{Andrew
  McStay}.} \bibinfo{year}{2018}\natexlab{}.
\newblock \showarticletitle{Fake news and the economy of emotions: Problems,
  causes, solutions}.
\newblock \bibinfo{journal}{\emph{Digital Journalism}}  \bibinfo{volume}{6}
  (\bibinfo{year}{2018}), \bibinfo{pages}{154--175}.
\newblock


\bibitem[\protect\citeauthoryear{Baly, Karadzhov, Alexandrov, Glass, and
  Nakov}{Baly et~al\mbox{.}}{2018}]%
        {D18-1389}
\bibfield{author}{\bibinfo{person}{Ramy Baly}, \bibinfo{person}{Georgi
  Karadzhov}, \bibinfo{person}{Dimitar Alexandrov}, \bibinfo{person}{James
  Glass}, {and} \bibinfo{person}{Preslav Nakov}.}
  \bibinfo{year}{2018}\natexlab{}.
\newblock \showarticletitle{Predicting Factuality of Reporting and Bias of News
  Media Sources}. In \bibinfo{booktitle}{\emph{Proc. of the Conf. on Empirical
  Methods in NLP}}. \bibinfo{pages}{3528--3539}.
\newblock


\bibitem[\protect\citeauthoryear{Barr{\'o}n-Cede{\~n}o, Martino, Jaradat, and
  Nakov}{Barr{\'o}n-Cede{\~n}o et~al\mbox{.}}{2019}]%
        {barron2019}
\bibfield{author}{\bibinfo{person}{Alberto Barr{\'o}n-Cede{\~n}o},
  \bibinfo{person}{Giovanni Da~San Martino}, \bibinfo{person}{Israa Jaradat},
  {and} \bibinfo{person}{Preslav Nakov}.} \bibinfo{year}{2019}\natexlab{}.
\newblock \showarticletitle{Proppy: A System to Unmask Propaganda in Online
  News}. In \bibinfo{booktitle}{\emph{Proc. of the 33th AAAI Conf. on
  Artificial Intelligence}}.
\newblock


\bibitem[\protect\citeauthoryear{Bird and Loper}{Bird and Loper}{2004}]%
        {bird2004nltk}
\bibfield{author}{\bibinfo{person}{Steven Bird} {and} \bibinfo{person}{Edward
  Loper}.} \bibinfo{year}{2004}\natexlab{}.
\newblock \showarticletitle{NLTK: the natural language toolkit}. In
  \bibinfo{booktitle}{\emph{Proc. of the Association for Computational
  Linguistics on Interactive poster and demonstration sessions}}.
  \bibinfo{pages}{31}.
\newblock


\bibitem[\protect\citeauthoryear{Fairbanks, Fitch, Knauf, and
  Briscoe}{Fairbanks et~al\mbox{.}}{2018}]%
        {fairbanks2018credibility}
\bibfield{author}{\bibinfo{person}{James Fairbanks}, \bibinfo{person}{Natalie
  Fitch}, \bibinfo{person}{Nathan Knauf}, {and} \bibinfo{person}{Erica
  Briscoe}.} \bibinfo{year}{2018}\natexlab{}.
\newblock \showarticletitle{Credibility Assessment in the News: Do we need to
  read?}. In \bibinfo{booktitle}{\emph{Proc. of the MIS2 Workshop held in
  conjuction with 11th Int'l Conf. on Web Search and Data Mining}}.
  \bibinfo{pages}{799--800}.
\newblock


\bibitem[\protect\citeauthoryear{Geurts, Ernst, and Wehenkel}{Geurts
  et~al\mbox{.}}{2006}]%
        {geurts2006extremely}
\bibfield{author}{\bibinfo{person}{Pierre Geurts}, \bibinfo{person}{Damien
  Ernst}, {and} \bibinfo{person}{Louis Wehenkel}.}
  \bibinfo{year}{2006}\natexlab{}.
\newblock \showarticletitle{Extremely randomized trees}.
\newblock \bibinfo{journal}{\emph{Machine learning}}  \bibinfo{volume}{63}
  (\bibinfo{year}{2006}), \bibinfo{pages}{3--42}.
\newblock


\bibitem[\protect\citeauthoryear{Gottfried and Shearer}{Gottfried and
  Shearer}{2016}]%
        {pew2016}
\bibfield{author}{\bibinfo{person}{Jeffrey Gottfried} {and}
  \bibinfo{person}{Elisa Shearer}.} \bibinfo{year}{2016}\natexlab{}.
\newblock \bibinfo{title}{News Use across Social Media Platforms 2016}.
\newblock
  \bibinfo{howpublished}{\url{http://www.journalism.org/2016/05/26/news-use-across-social-media-platforms-2016}}.
\newblock


\bibitem[\protect\citeauthoryear{Horne and Adali}{Horne and Adali}{2017}]%
        {horne2017just}
\bibfield{author}{\bibinfo{person}{Benjamin~D Horne} {and}
  \bibinfo{person}{Sibel Adali}.} \bibinfo{year}{2017}\natexlab{}.
\newblock \showarticletitle{This just in: fake news packs a lot in title, uses
  simpler, repetitive content in text body, more similar to satire than real
  news}. In \bibinfo{booktitle}{\emph{Proc. of the 2nd Intn'l Workshop on News
  and Public Opinion}}.
\newblock


\bibitem[\protect\citeauthoryear{Kraskov, St{\"o}gbauer, and
  Grassberger}{Kraskov et~al\mbox{.}}{2004}]%
        {kraskov2004estimating}
\bibfield{author}{\bibinfo{person}{Alexander Kraskov}, \bibinfo{person}{Harald
  St{\"o}gbauer}, {and} \bibinfo{person}{Peter Grassberger}.}
  \bibinfo{year}{2004}\natexlab{}.
\newblock \showarticletitle{Estimating mutual information}.
\newblock \bibinfo{journal}{\emph{Physical review E}}  \bibinfo{volume}{69}
  (\bibinfo{year}{2004}), \bibinfo{pages}{066138}.
\newblock


\bibitem[\protect\citeauthoryear{Lazer, Baum, Benkler, Berinsky, Greenhill,
  Menczer, Metzger, Nyhan, Pennycook, Rothschild, et~al\mbox{.}}{Lazer
  et~al\mbox{.}}{2018}]%
        {lazer2018science}
\bibfield{author}{\bibinfo{person}{David~MJ Lazer}, \bibinfo{person}{Matthew~A
  Baum}, \bibinfo{person}{Yochai Benkler}, \bibinfo{person}{Adam~J Berinsky},
  \bibinfo{person}{Kelly~M Greenhill}, \bibinfo{person}{Filippo Menczer},
  \bibinfo{person}{Miriam~J Metzger}, \bibinfo{person}{Brendan Nyhan},
  \bibinfo{person}{Gordon Pennycook}, \bibinfo{person}{David Rothschild},
  {et~al\mbox{.}}} \bibinfo{year}{2018}\natexlab{}.
\newblock \showarticletitle{The science of fake news}.
\newblock \bibinfo{journal}{\emph{Science}} (\bibinfo{year}{2018}).
\newblock


\bibitem[\protect\citeauthoryear{Lesne}{Lesne}{2014}]%
        {lesne2014shannon}
\bibfield{author}{\bibinfo{person}{Annick Lesne}.}
  \bibinfo{year}{2014}\natexlab{}.
\newblock \showarticletitle{Shannon Entropy: A Rigorous Notion at The
  Crossroads Between Probability, Information Theory, Dynamical Systems and
  Statistical Physics}.
\newblock \bibinfo{journal}{\emph{Mathematical Structures in Computer Science}}
   \bibinfo{volume}{24} (\bibinfo{year}{2014}), \bibinfo{pages}{240--311}.
\newblock


\bibitem[\protect\citeauthoryear{Pennebaker, Boyd, Jordan, and
  Blackburn}{Pennebaker et~al\mbox{.}}{2015}]%
        {pennebaker2015liwc}
\bibfield{author}{\bibinfo{person}{James Pennebaker}, \bibinfo{person}{Ryan
  Boyd}, \bibinfo{person}{Kayla Jordan}, {and} \bibinfo{person}{Kate
  Blackburn}.} \bibinfo{year}{2015}\natexlab{}.
\newblock \showarticletitle{The Development and Psychometric Properties of
  LIWC2015}. 10.15781/T29G6Z.
\newblock


\bibitem[\protect\citeauthoryear{P{\'e}rez-Rosas, Kleinberg, Lefevre, and
  Mihalcea}{P{\'e}rez-Rosas et~al\mbox{.}}{2018}]%
        {perez@coling2018}
\bibfield{author}{\bibinfo{person}{Ver{\'o}nica P{\'e}rez-Rosas},
  \bibinfo{person}{Bennett Kleinberg}, \bibinfo{person}{Alexandra Lefevre},
  {and} \bibinfo{person}{Rada Mihalcea}.} \bibinfo{year}{2018}\natexlab{}.
\newblock \showarticletitle{Automatic Detection of Fake News}. In
  \bibinfo{booktitle}{\emph{Proc. of Int'l Conf. on Computational
  Linguistics}}. \bibinfo{pages}{3391--3401}.
\newblock


\bibitem[\protect\citeauthoryear{Pham}{Pham}{[n. d.]}]%
        {buzzfeed2017git}
\bibfield{author}{\bibinfo{person}{Scott Pham}.} \bibinfo{year}{[n.
  d.]}\natexlab{}.
\newblock \bibinfo{title}{Analysis of fake news sites and viral posts, 2016 vs.
  2017}.
\newblock
  \bibinfo{howpublished}{\url{https://github.com/BuzzFeedNews/2017-12-fake-news-top-50}}.
\newblock


\bibitem[\protect\citeauthoryear{Potthast, Kiesel, Reinartz, Bevendorff, and
  Stein}{Potthast et~al\mbox{.}}{2017}]%
        {potthast@arxiv2017}
\bibfield{author}{\bibinfo{person}{Martin Potthast}, \bibinfo{person}{Johannes
  Kiesel}, \bibinfo{person}{Kevin Reinartz}, \bibinfo{person}{Janek
  Bevendorff}, {and} \bibinfo{person}{Benno Stein}.}
  \bibinfo{year}{2017}\natexlab{}.
\newblock \showarticletitle{A Stylometric Inquiry into Hyperpartisan and Fake
  News}.
\newblock \bibinfo{journal}{\emph{CoRR}}  \bibinfo{volume}{abs/1702.05638}
  (\bibinfo{year}{2017}).
\newblock


\bibitem[\protect\citeauthoryear{Rashkin, Choi, Jang, Volkova, and
  Choi}{Rashkin et~al\mbox{.}}{2017}]%
        {rashkin2017truth}
\bibfield{author}{\bibinfo{person}{Hannah Rashkin}, \bibinfo{person}{Eunsol
  Choi}, \bibinfo{person}{Jin~Yea Jang}, \bibinfo{person}{Svitlana Volkova},
  {and} \bibinfo{person}{Yejin Choi}.} \bibinfo{year}{2017}\natexlab{}.
\newblock \showarticletitle{Truth of varying shades: Analyzing language in fake
  news and political fact-checking}. In \bibinfo{booktitle}{\emph{Proc. of the
  Conf. on Empirical Methods in NLP}}. \bibinfo{pages}{2931--2937}.
\newblock


\bibitem[\protect\citeauthoryear{Shu, Sliva, Wang, Tang, and Liu}{Shu
  et~al\mbox{.}}{2017a}]%
        {shu2017fake}
\bibfield{author}{\bibinfo{person}{Kai Shu}, \bibinfo{person}{Amy Sliva},
  \bibinfo{person}{Suhang Wang}, \bibinfo{person}{Jiliang Tang}, {and}
  \bibinfo{person}{Huan Liu}.} \bibinfo{year}{2017}\natexlab{a}.
\newblock \showarticletitle{Fake News Detection on Social Media: A Data Mining
  Perspective}.
\newblock \bibinfo{journal}{\emph{ACM SIGKDD Explorations Newsletter}}
  \bibinfo{volume}{19} (\bibinfo{year}{2017}), \bibinfo{pages}{22--36}.
\newblock


\bibitem[\protect\citeauthoryear{Shu, Wang, and Liu}{Shu
  et~al\mbox{.}}{2017b}]%
        {shu2017exploiting}
\bibfield{author}{\bibinfo{person}{Kai Shu}, \bibinfo{person}{Suhang Wang},
  {and} \bibinfo{person}{Huan Liu}.} \bibinfo{year}{2017}\natexlab{b}.
\newblock \showarticletitle{Exploiting Tri-Relationship for Fake News
  Detection}.
\newblock \bibinfo{journal}{\emph{CoRR}}  \bibinfo{volume}{abs1712.07709}
  (\bibinfo{year}{2017}).
\newblock


\bibitem[\protect\citeauthoryear{Silverman}{Silverman}{2015}]%
        {silverman2015lies}
\bibfield{author}{\bibinfo{person}{Craig Silverman}.}
  \bibinfo{year}{2015}\natexlab{}.
\newblock \showarticletitle{Lies, damn lies, and viral content: How news
  websites spread (and Debunk) online rumors, unverified claims and
  misinformation}.
\newblock \bibinfo{journal}{\emph{Tow Center for Digital Journalism}}
  \bibinfo{volume}{168} (\bibinfo{year}{2015}).
\newblock


\bibitem[\protect\citeauthoryear{Silverman and Singer-Vine}{Silverman and
  Singer-Vine}{2016}]%
        {silverman2016}
\bibfield{author}{\bibinfo{person}{Craig Silverman} {and}
  \bibinfo{person}{Jeremy Singer-Vine}.} \bibinfo{year}{2016}\natexlab{}.
\newblock \bibinfo{title}{Most Americans Who See Fake News Believe It, New
  Survey Says}.
\newblock
\newblock
\newblock
\shownote{BuzzFeed News.}


\bibitem[\protect\citeauthoryear{Wu and Liu}{Wu and Liu}{2018}]%
        {wu2018tracing}
\bibfield{author}{\bibinfo{person}{Liang Wu} {and} \bibinfo{person}{Huan Liu}.}
  \bibinfo{year}{2018}\natexlab{}.
\newblock \showarticletitle{Tracing fake-news footprints: Characterizing social
  media messages by how they propagate}. In \bibinfo{booktitle}{\emph{Proc. of
  the 11th Int'l Conf. on Web Search and Data Mining}}.
  \bibinfo{pages}{637--645}.
\newblock


\bibitem[\protect\citeauthoryear{Zhao and Yu}{Zhao and Yu}{2006}]%
        {zhao2006model}
\bibfield{author}{\bibinfo{person}{Peng Zhao} {and} \bibinfo{person}{Bin Yu}.}
  \bibinfo{year}{2006}\natexlab{}.
\newblock \showarticletitle{On model selection consistency of Lasso}.
\newblock \bibinfo{journal}{\emph{Journal of Machine learning research}}
  \bibinfo{volume}{7} (\bibinfo{year}{2006}), \bibinfo{pages}{2541--2563}.
\newblock


\end{thebibliography}

\end{document}